\documentclass[journal]{IEEEtran}
\usepackage{amsmath,amssymb,amsfonts}
\usepackage{algorithmic}
\usepackage{array}
\usepackage[caption=false,font=normalsize,labelfont=sf,textfont=sf]{subfig}
\usepackage{stfloats}
\usepackage{url}
\usepackage{graphicx}
\usepackage{xcolor}      
\usepackage{hyperref}    
\usepackage{multirow}
\usepackage{enumitem}
\usepackage{arydshln}
\usepackage[T1]{fontenc}
\newtheorem{remark}{Remark}
\usepackage[numbers,sort&compress]{natbib}

\definecolor{mypink}{RGB}{255, 105, 180}  

\hypersetup{
    colorlinks=true,  
    urlcolor=mypink,  
}

\def\BibTeX{{\rm B\kern-.05em{\sc i\kern-.025em b}\kern-.08em
    T\kern-.1667em\lower.7ex\hbox{E}\kern-.125emX}}
\usepackage{balance}
\begin{document}
\title{Multi-Level Embedding and Alignment Network with Consistency and Invariance Learning for Cross-View Geo-Localization} 
\author{Zhongwei Chen, Zhao-Xu Yang, Hai-Jun Rong\\
\thanks{This paper is submitted for review on December 23, 2024. This work was supported in part by the Key Research and Development Program of Shaanxi, PR China (No. 2023-YGBY-235), the National Natural Science Foundation of China (No. 61976172 and No. 12002254), Major Scientific and Technological Innovation Project of Xianyang, PR China (No. L2023-ZDKJ-JSGG-GY-018). (Corresponding author: Zhao-Xu Yang and Hai-Jun Rong)}

\thanks{Zhongwei Chen, Zhao-Xu Yang and Hai-Jun Rong  are with the State Key Laboratory for Strength and Vibration of Mechanical Structures, Shaanxi Key Laboratory of Environment and Control for Flight Vehicle, School of Aerospace Engineering, Xi’an Jiaotong University, Xi’an 710049, PR China (e-mail:ISChenawei@stu.xjtu.edu.cn; yangzhx@xjtu.edu.cn; hjrong@mail.xjtu.edu.cn).}}

\maketitle
\begin{abstract}
Cross-View Geo-Localization (CVGL) involves determining the localization of aerial images by retrieving the most similar GPS-tagged satellite images. However, the imaging gaps between platforms are often significant and the variations in viewpoints are substantial, which limits the ability of existing methods to effectively associate cross-view features and extract consistent and invariant characteristics. Moreover, existing methods often overlook the problem of increased computational and storage requirements when improving model performance. To handle these limitations, we propose a lightweight enhanced alignment network, called the multi-level embedding and alignment network (MEAN). The MEAN framework uses a progressive multi-level enhancement strategy, global-to-local associations, and cross-domain alignment, enabling feature communication across levels. This allows MEAN to effectively connect features at different levels and learn robust cross-view consistent mappings and cross-view invariant features. Moreover, MEAN adopts a shallow backbone network combined with a lightweight branch design, effectively reducing parameter count and computational complexity. Experimental results on the University-1652 and SUES-200 datasets demonstrate that MEAN reduces parameter count by 62.17\% and computational complexity by 70.99\% compared with state-of-the-art models, while maintaining competitive or even superior performance. 
{Our code is available at {\href{https://github.com/ISChenawei/MEAN}{https://github.com/ISChenawei/MEAN}}} 

\end{abstract}

\section{Introduction}
\IEEEPARstart{C}{ross-view} geo-localization (CVGL) has received extensive attention in autonomous vehicles, aerial photography, and autonomous navigation \cite{zheng2020university,long2017accurate,humenberger2022investigating}. CVGL is usually recognized as an image retrieval task on the heterogeneous platform to accurately determine the geo-location of a query image by matching it with several reference images captured from the varying views of different platforms. The early applications were concentrated on the matching of ground panoramic images with satellite images \cite{zhang2023cross,deuser2023sample4geo,zhang2024geodtr+}. {\color{black}In recent years, with the continuous advancement of remote sensing technology \cite{qin2024fdgnet,qin2024cross,feng2024cross}, the applications of drone aerial photography have been further expanded, encompassing drone target localization, drone navigation, and other related fields \cite{shen2023mccg,ge2024multibranch,chen2024sdpl}}. In these cases, the aerial images captured by drones can be matched with satellite images of the same geographic area with precise longitude and latitude coordinates, achieving indirect drone location. This realizes drone navigation and replaces the need for global navigation satellite system (GNSS) equipment onboard \cite{zhao2024transfg}. However, CVGL encounters substantial challenges due to appearance variations and spatial misalignment caused by imaging gaps, scale variations and spatial transformations in the aerial photography scene.

Previous research for CVGL focused on improving feature representation capabilities, initially using global features to encompass overall structural information \cite{ding2020practical,lin2022joint} or local features to detail fine-grained aspects \cite{wang2021each,hu2022beyond}. However, global features lack the ability to represent detailed information, whereas local features are sensitive to variations in viewpoint or scale. Consequently, some research has developed the joint global-local feature representations to overcome their respective inherent limitations with the more powerful feature extraction architectures. For example, the Transformer \cite{yang2021cross,zhu2022transgeo} was used to model long-range dependencies and complex contextual semantics at both the global and local levels. Although these architectures possess strong feature extraction capabilities to enhance overall feature representation and mitigate the discrepancies caused by appearance variations, the disparity in feature space distribution due to spatial misalignment makes it difficult to obtain efficient alignment of cross-view features, namely feature consistency. Moreover, the excessive emphasis on contextual information can induce noise interference \cite{ge2024multi} to affect the consistency of the characteristics and the accuracy of matching in complex retrieval tasks. Furthermore, increasing architecture complexity results in a high parameter count and significantly imposes a higher computational load on limited airborne resources \cite{mcenroe2022survey}.

Attempts to extract the cross-view consistent features provided the possibilities to accurately align and associate the same target from different viewpoints and transformed spatial \cite{xia2024enhancing}. Several methods \cite{lin2022joint,shi2020optimal,sun2023f3} have been proposed to address inconsistencies in the feature distribution. These methods combined dense partition learning \cite{chen2024sdpl,dai2021transformer} and feature alignment strategies \cite{tian2021uav,xia2024enhancing} to capture deep structural relationships of cross-view features within a shared embedding space. To further improve the discriminability of features in the cross-view mappings of these methods, contrastive loss \cite{deuser2023sample4geo} and triplet loss \cite{2019Lending} have been used to achieve a more precise differentiation between positive and negative samples within the embedding space.

Although these methods have made significant progress in addressing feature distribution inconsistencies through dense partition learning and feature alignment, they often overemphasized fine-grained feature alignment. In this case, it relied heavily on the details of the viewpoints or scales, making it difficult to effectively associate features across different levels. In scenarios characterized by substantial viewpoint differences or large-scale spatial variations, consistency mappings may not capture the critical cross-view invariant that differs from these environmentally sensitive features, i.e., the invariance of features \cite{cui2021cross}. This limited their robustness and uniformity, ultimately undermining the generalization capability.

\begin{figure}[t]
  \centering
  \includegraphics[width=3.9in]{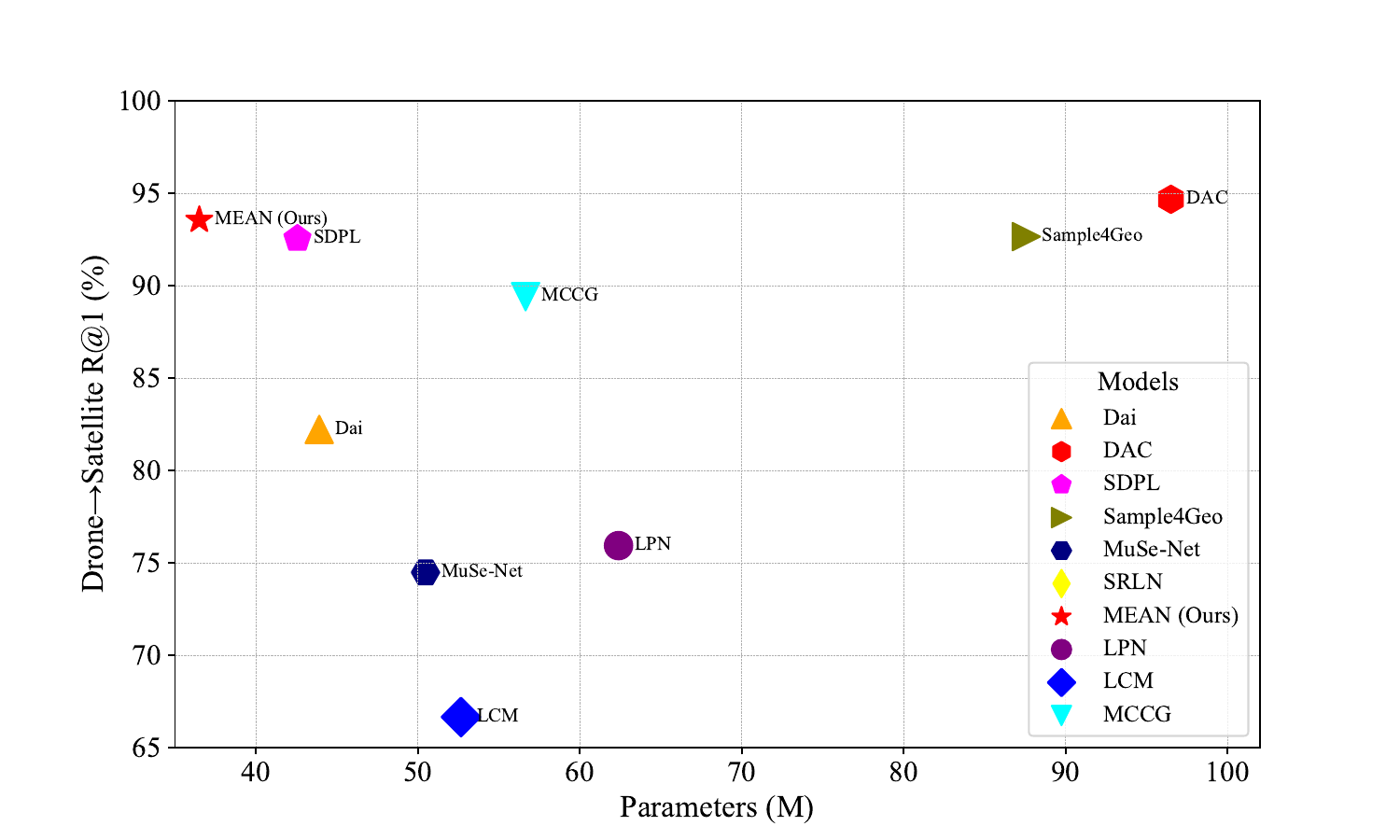}
  \caption{The balance between model performance and parameter count. Model performance is evaluated based on the R@1 accuracy on the Drone$\rightarrow$Satellite from the University-1652 dataset. Our method achieves superior performance with a lower parameter count compared with state-of-the-art (SOTA) methods, demonstrating efficiency in CVGL tasks.}
  \label{fig1}
  \end{figure}

To address those issues, we propose a novel enhanced alignment network for the CVGL tasks, called multi-level embedding alignment network (MEAN). In the proposed MEAN, ConvNeXt-Tiny, the smallest network in the ConvNeXt family \cite{liu2022convnet} is used as a backbone to extract coarse-grained features. These features are then processed separately by the three designed branches. The progressive diversification embedding branch proposed in the MEAN adopts a stepwise expansion strategy to generate a rich set of embedding features, allowing the model to learn diverse feature representations. Furthermore, it uses contrastive loss to enhance the consistency and discriminability of the shared embeddings. To better coordinate feature representation capabilities, the other branch, named the global expansion embedding branch, jointly optimizes global and fine-grained feature representations to realize the global-to-local associations. Additionally, the multi-level feature fusion and adaptive calibration strategies with multi-level constraint achieve precise and robust feature alignment in the embedding space in the cross-domain enhanced alignment branch. Therefore, the proposed MEAN not only enhances feature representation, but also deeply explores the invariant modes and potential commonalities within the features. As shown in Fig. \ref{fig1}, our method significantly achieves improvements in both matching accuracy and computational efficiency compared with existing state-of-the-art methods. Its main contributions are summarized as follows:

\begin{itemize}
  \item The proposed MEAN establishes a joint lightweight learning framework that possesses the shallow feature representation and deep latent structure learning to incorporate multi-level characteristics and semantic depth. It enhances the feature representation capacity through collaborative embedding branches. Moreover, the consistency and invariance of cross-view features are ensured through the multi-level feature fusion and adaptive calibration strategies with multi-level constraints.
  \item We propose two embedding branches to facilitate the collaborative modeling of global and fine-grained features. The contrastive loss-based stepwise expansion strategy is utilized in the progressive expansion branch to incrementally generate diverse embedding representations, while the global expansion branch ensures the global-to-local associations and structural integrity of the feature space.
  \item In the cross-domain enhanced alignment branch, the multi-level feature fusion and adaptive calibration strategies with multi-level constraints realize the consistency learning and effectively ensure robust invariance under cross-domain transformations.
  \item Compared with state-of-the-art models, extensive experiments demonstrate that the proposed MEAN achieves a reduction of 62.17\% in parameter count and a decrease of 70.99\% in computational complexity while maintaining competitive or even superior performance.
\end{itemize}

{\color{black}The remainder of this paper is organized as follows. Section \ref{related works} systematically reviews prior research. In Section \ref{MEAN}, the proposed MEAN is presented in detail. The experimental results are reported and analyzed in Section \ref{results}. In Section  \ref{discussion}, a comprehensive analysis of the proposed MEAN is provided. Finally, conclusions are outlined in Section \ref{conclusions}.}
\section{RELATED WORKS}\label{related works}
{\color{black}In this section, we provide a concise review of the related work on CVGL, with an emphasis on two main methods commonly employed in CVGL tasks: 1) feature extraction and contextual enhancement, and 2) feature alignment and optimization for discriminative power.}

{\color{black}\subsection{Cross-View Geo-Localization}
CVGL has been generally regarded as an image retrieval task involving heterogeneous viewpoints, primarily focusing on ground-to-satellite and drone-to-satellite image retrieval. Early studies relied on handcrafted feature operators to extract and align features across distinct viewpoints \cite{Lin_2013_CVPR,6130498}. With the rapid advancement of deep convolutional neural networks (CNNs), recent works \cite{Cai_2019_ICCV,Shi_2020_CVPR,Toker_2021_CVPR,Qian_2023_ICCV} have shifted toward deep representation learning to learn cross-view features. This progress brought creation of multiple public datasets featuring ground–satellite image pairs such as CVUSA \cite{Zhai_2017_CVPR}, CVACT \cite{Liu_2019_CVPR}. A pre-trained CNN for high-level feature extraction in CVGL was first employed to demonstrate that the features encoded the semantic cues related to geographic location \cite{Workman_2015_CVPR_Workshops}. Subsequently, NetVLAD \cite{Arandjelovic_2016_CVPR} was incorporated into a Siamese-like architecture to achieve robust image descriptors with significant viewpoint variation \cite{Zhai_2017_CVPR}. 
    
CVGL has been further extended to the drone-view, along with several specialized drone-view localization datasets. The University-1652 dataset has contained drone–satellite image pairs and reformulated the retrieval problem within a classification framework \cite{zheng2020university}. Subsequent studies expanded on this dataset by incorporating diverse environmental conditions, such as fog, rain, and snow, to systematically assess model robustness under real-world scenarios \cite{wang2024multiple}. Additionally, the SUES-200 dataset \cite{zhu2023sues} was proposed to specifically examine the influence of varying flight altitudes on CVGL performance.}

\subsection{Feature Extraction and Context Enhancement}
The first category of methods focused primarily on enhancing feature representation, typically achieving CVGL through single global or local feature extraction. LCM \cite{ding2020practical} mapped aerial and satellite images into a unified feature space, transforming the task into a global location classification to learn the overall structure of the image. However, the ability to learn fine-grained features was limited in such global methods. Therefore, LPN \cite{wang2021each} introduced a local pattern partitioning method that segmented images into multiple distance-based regions using a square ring partitioning strategy. This method effectively extracted fine-grained contextual information surrounding the target. Although this local method improved the model’s sensitivity to detail, it was found to be limited in handling significant viewpoint and scale variations by the extraction of the single global or local feature. Moreover, relying on a single branch for processing either global or local features hindered the effective integration of multi-scale and multi-level spatial information in cross-view scenarios. To address these issues, IFSs \cite{ge2024multibranch} proposed a multi-branch joint representation learning strategy that integrated global and local feature information through a multi-branch structure.

However, these methods were mainly based on convolutional neural networks (CNNs), which are inherently limited in modeling long-range dependencies and complex contextual semantics, particularly in CVGL tasks \cite{dai2021transformer}. Transformer and ConvNeXt architectures have been gradually adopted in CVGL due to their advantages in long-range dependency modeling and contextual information extraction. TransFG \cite{zhao2024transfg} used Transformer-based feature aggregation and gradient-guided modules to effectively integrate global and local information. SRLN \cite{lv2024direction} based on Swin Transformer \cite{liu2021swin} combined direction guidance and multi-scale feature fusion strategies, effectively bridging viewpoint and scale discrepancies. MCCG \cite{shen2023mccg}, on the contrary, introduced a multi-classifier method based on ConvNeXt to learn rich feature representations. However, the emphasis on contextual information can lead to over-attention to non-essential features. To mitigate this, CCR \cite{du2024ccr} introduced counterfactual causal reasoning to strengthen the model’s attention mechanism, which is useful to distinguish between essential and non-essential features. MFJR \cite{ge2024multi} used a multi-level feedback joint representation learning method, incorporating an adaptive region elimination strategy to effectively suppress irrelevant information and focus on key target features.

Despite the effectiveness of Transformer-based and large ConvNeXt architectures with specific multi-level learning strategies in capturing features, their high computational costs pose challenges in resource-constrained environments. Furthermore, under significant viewpoint or scale variations, it is difficult to maintain the robustness and consistency in cross-view matching.

\subsection{Feature Alignment and Discriminative Optimization}

To achieve consistency in extracting features across different cross-views, researchers have utilized spatial alignment strategies to learn more consistent and discriminative features within a shared feature space. PCL \cite{tian2021uav} utilized perspective projection transformation to align aerial and satellite images to reduce spatial misalignment between views, and then employed a CGAN to synthesize realistic satellite image styles to narrow the imaging gap. This explicit alignment method struggled to learn internal feature differences effectively and might add unnecessary noise. Consequently, many researchers have attempted to integrate the metric loss and the contrastive loss within a shared space to identify the internal discrepancies. However, this method is often challenged by interference from viewpoint variations and exhibits limited discriminative power for negative samples. To address this, Sample4Geo \cite{deuser2023sample4geo} employed a hard negative sampling strategy to improve the model's feature discrimination and contrastive learning effectiveness. However, significant differences in feature representations across cross-views posed challenges to the extraction and alignment of geographical features under substantial viewpoint variations. Relying solely on constraint-based alignment within a shared embedding space has been shown to be insufficient to address these discrepancies. Therefore, CAMP \cite{wu2024camp} introduced contrastive attribute mining and position-aware partitioning strategies to align geographic features under varying viewpoints and scales. Although this method has enhanced local feature consistency and discrimination, it excelled in extracting explicit differences across viewpoints but struggled to learn feature invariance under significant changes in viewpoint and scale, thereby limiting the model’s generalization capability. DAC \cite{xia2024enhancing} adopted domain alignment and scene consistency constraints to achieve coarse-to-fine feature consistency, relying primarily on direct alignment without fully exploring deep invariant patterns across viewpoints. As a consequence, DAC encountered challenges in maintaining stable feature mappings under extreme variations in viewpoint or scale. Furthermore, both CAMP and DAC incurred high computational costs in enhancing alignment accuracy, thereby affecting model efficiency and limiting deployment flexibility.

\begin{figure*}[t]
  \centering
  \includegraphics[width=7in]{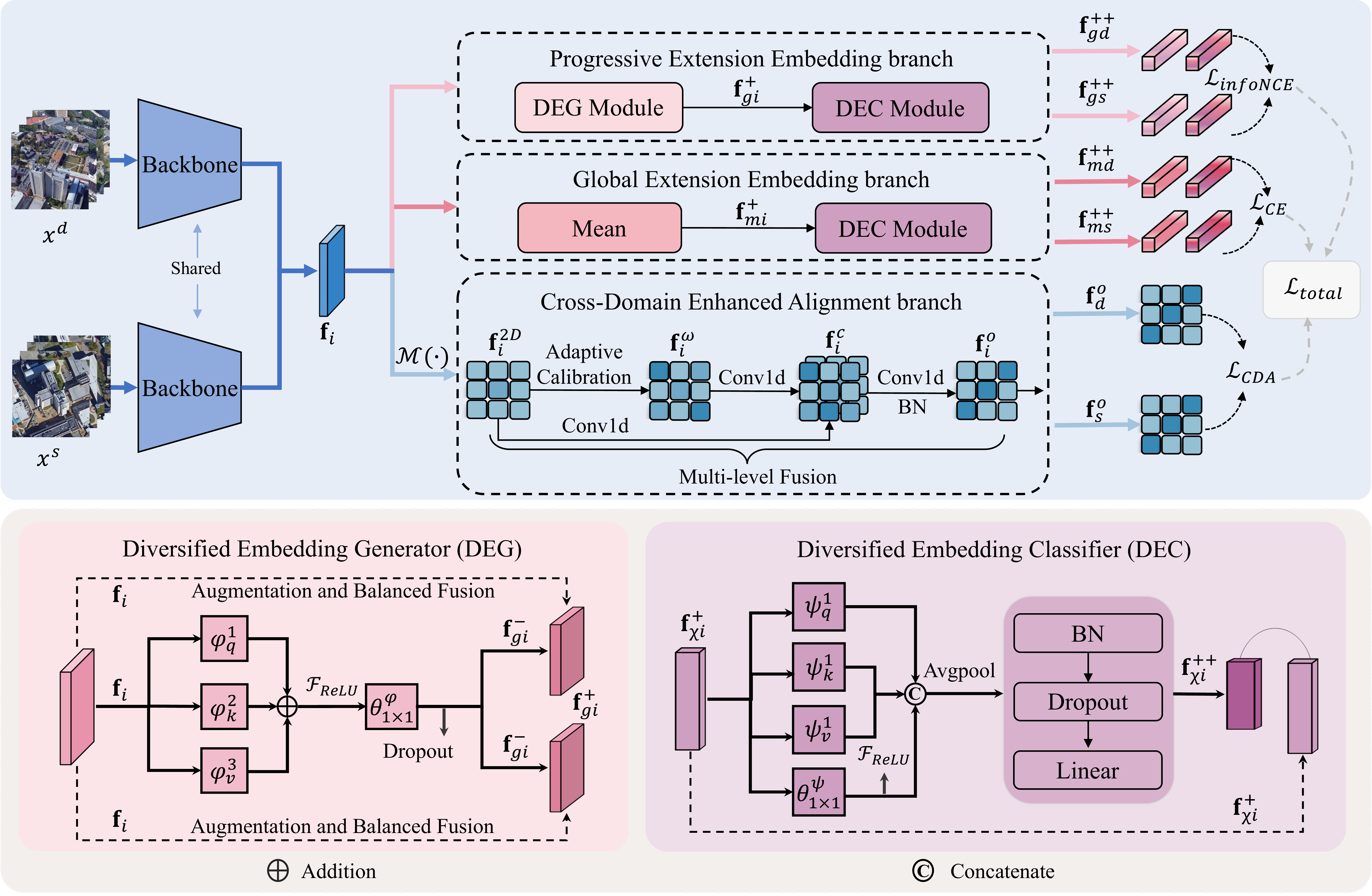}
  \caption{The pipeline of the proposed network includes a ConvNeXt-Tiny backbone and three core branches. The progressive extension embedding branch (PEE) learns multi-scale embedding features through progressive multi-scale convolutions optimized by the loss \(\mathcal{L}_{\text{InfoNCE}}\) to enhance diverse feature representations and discriminative ability. The global extension embedding branch (GEE) aggregates global and locally generated embedding features optimized by the loss \( \mathcal{L}_{\text{CE}} \). The cross-domain enhanced alignment branch (CEA) uses multi-level fusion and adaptive calibration strategy with a novel loss \(\mathcal{L}_{\text{CDA}}\) to dynamically adjust feature consistency within a shared latent space of high-dimensional embeddings. For simplicity, let $i\in\{d,s\}$ denote drone view ($d$) and satellite view ($s$), and $\chi\in\{g,m\}$ represent diversified embedding generator (DEG) module ($g$) and Mean ($m$).
}
  \label{fig2}  
  \end{figure*}
\section{Multi-Level Embedding and Alignment Network}\label{MEAN}

The proposed MEAN framework is introduced in this section, with an overview of this model illustrated in Fig.\ref{fig2}. MEAN utilizes ConvNeXt-Tiny as the backbone to extract initial coarse-grained features. The progressive extension embedding (PEE) branch and global extension embedding (GEE) branch generate multi-embedding representations, seizing representative cross-view feature cues, and refining global spatial representations, respectively. Additionally, the cross-domain enhanced alignment (CEA) branch is integrated to exhaustively extract underlying structural patterns across different view images. During the training phase, distinct loss functions are utilized for the features extracted from each branch. In particular, a cross-joint optimization strategy is employed in the PEE and CEA branches. By incorporating multi-level constraints, MEAN promotes feature enhancement and alignment, while optimizing the adaptability of feature representation for cross-view consistency and modality invariance.

\textit{Problem Formulation}: Given a CVGL dataset, we denote the input image pairs as $\{x^d, x^s\}$, where $x^d$ and $x^s$ represent the images captured from drone and satellite views, respectively. Each image is associated with a label $y =1,\cdots,C$, where $C$ denotes the total number of categories corresponding to different locations in the dataset. For example, in the University-1652 dataset, there are 701 buildings, each containing one satellite view image and multiple drone view images. For the CVGL task, the goal is to learn a mapping function that projects images from different platforms (e.g., drone and satellite) into a shared semantic space. In this space, the features of images representing the same location should be close to each other, while those from different locations should be well separated. This method enables effective matching of images from different views in the same location by leveraging feature similarity. It can address the challenges of significant viewpoint differences and supports accurate geo-localization.
\subsection{ConvNeXt-Tiny Backbone for Feature Extraction}
In this work, we use ConvNeXt-Tiny as the backbone for the extraction of features. ConvNeXt-Tiny is the most lightweight variant within the ConvNeXt family, combining depthwise convolutions and layer normalization to provide an efficient CNN solution for CVGL tasks.

ConvNeXt-Tiny processes input images with a resolution of 384×384 pixels, and the output feature maps are represented as follows:
\begin{equation}
\mathbf{f}_i = \mathcal{F}_{\text{backbone}}(x^i), \quad i \in \{d, s\}
\end{equation}
where \(x^i\) represents input images from drone views \(\{x^d\}\) and satellite  views \(\{x^s\}\), and \(\mathcal{F}_{\text{backbone}}\) denotes the feature extraction function of ConvNeXt-Tiny. The extracted feature maps \( \mathbf{f}_i \in \mathbb{R}^{C_i \times H_i \times W_i}\) capture the underlying hierarchical patterns that are critical for subsequent procedures, where \( C_i \), \( H_i \) and \( W_i \) represent the number of channels, height and width, respectively. Additionally, weight sharing is implemented between drone and satellite branches, which not only enhances cross-domain feature consistency but also reduces computational overhead. 

{\remark{Compared with deeper and more complex backbone networks such as Transformer-based \cite{ge2024multi} and ConvNeXt-based \cite{du2024ccr} models for CVGL, ConvNeXt-Tiny reduces the parameter count by 67\%. However, because of the shallow structure and compact design, it has certain limitations to learning complex and deep information about features. To address these shortcomings, we introduce complementary branches, designed as follows.}}
\subsection{Progressive Extension Embedding branch}
The proposed progressive extension embedding branch (PEE) utilizes a multi-branch convolutional structure to fuse information across different scales. It enhances feature embeddings and enables precise learning of complex semantic patterns in multi-scale scenarios. The PEE branch consists of two sub-modules, namely the diversified embedding generator (DEG) and the diversified embedding classifier (DEC).

{\it DEG module}: DEG extracts contextual information from multiple embeddings through parallel pathways, enhancing representational consistency in different embeddings. As illustrated in Fig.\ref{fig2}, for the output of the feature map by the backbone network \(\mathbf{f}_i\in \mathbb{R}^{ C_i \times H_i \times W_i} \), three dilated convolution layers with a kernel size of \({3 \times 3} \) are applied. These layers, namely \( \varphi^1_q \), \( \varphi^2_k \), and \( \varphi^3_v\) have dilation rates of 1, 2 and 3, respectively. They reduce the channel size of the feature map to one-fourth of its original size. The feature maps are then fused by combining the outputs into a single feature map, followed by the ReLU activation layer \( F_\text{ReLU} \) to improve the nonlinearity of the DEG's representation. Subsequently, another convolutional layer with a kernel size of \( 1 \times 1 \), denoted as \( \theta^\varphi_{1 \times 1} \), is applied to the resulting feature map to restore its dimension to match that of the original feature map \( \mathbf{f}_i\). Based on this, we obtain compact and enhanced embedded features with improved diversity  \(\mathbf{f}_{gi}^{-}\). To adaptively address feature alignment and enhancement across different task scenarios, we introduce a feature augmentation and balanced fusion strategy. This strategy combines the generated embedding features \(\mathbf{f}_{gi}^{-}\) with the original features \( \mathbf{f}_i\) through weighted fusion, preserving essential information from the original features while enhancing the overall feature representation. Finally, the DEG module generates two identical feature representations, both of which undergo a dropout operation  \( \mathcal{D}(\cdot)\)  to mitigate overfitting and enhance the model generalization capability. These two sets of feature outputs will be used in subsequent classification procedure to expand the discriminative power of the model and enable it to co-optimize across different feature subspaces. The feature embedding \( \mathbf{f}_i^{+} \) is represented as follows:
\begin{equation}
\mathbf{f}_{gi}^{-} = \mathcal{D}(\theta^\varphi_{1 \times 1} ( F_\text{ReLU}( \frac{1}{3} ( \varphi^1_q (\mathbf{f}_i) + \varphi^2_k (\mathbf{f}_i) + \varphi^3_v (\mathbf{f}_{i}) ) )))
\end{equation}
\begin{equation}
\mathbf{f}_{gi}^{+} = \mathbf{\omega} \left( \mathbf{f}_{gi}^{-} + \mathbf{f}_i \right)
\end{equation}
where \( \mathbf{\omega} \) serves as a augmentation and balanced fusion factor to weight the contributions of \( \mathbf{f}_{gi}^{-} \) and \( \mathbf{f}_i \), and the collective set of generated embeddings \( \mathbf{f}_{gi}^{+}\) is subsequently employed as the input for the DEC module.

{\it DEC module}\label{DEC module}: To address the potential oversight of fine-grained details and cross-scale contextual information in previous diversified embeddings, the DEC module adopts a broader receptive field to obtain diverse embedded local details. In addition, it incorporates a progressive feature enhancement strategy to further explore and enhance the representational capacity of the embeddings. Specifically, we use four branches, three of which apply dilated convolution layers \( \psi^1_q \), \( \psi^2_k \) and \( \psi^3_v \) with dilation rates of 1, 2 and 3 corresponding to different scales. These convolution layers expand the receptive field of the network, enabling the model to learn both fine details and wide contextual information without increasing the number of parameters. Furthermore, we introduce an extra \( 1 \times 1 \) convolution \(\theta^{\psi}_{1 \times 1} \) to further adjust the embeddings distribution and improve the balance between local and global embeddings representations. Therefore, the generated features are concatenated \({C}(\cdot)\) and fused to form a unified embeddings representation.

Subsequently, adaptive average pooling (AvgPool) is applied to aggregate the multi-scale information, producing more discriminative semantic embeddings. Finally, the compact feature representation is processed through the batch normalization operation \( \mathcal{B}(\cdot)\), followed by a dropout operation \( {D}(\cdot)\) for regularization to avoid overfitting, and then propagated to the linear layer \( {L}(\cdot)\) for classification. We outline these processing operations \({C}(\cdot)\), \( \mathcal{B}(\cdot)\), \( {D}(\cdot)\), \( {L}(\cdot)\) as a lump \( \mathcal{P}(\cdot)\):
\begin{equation}\label{eq04}
\mathbf{f}_{gi}^{++} = \mathcal{P}(\psi^1_q (\mathbf{f}_{gi}^{+}), \psi^2_k (\mathbf{f}_{gi}^{+}), \psi^3_v (\mathbf{f}_{gi}^{+}), \theta^{\psi}_{1\times1} (\mathbf{f}_{gi}^{+}))
\end{equation}

{\remark{Compared with the method \cite{ge2024multi} that extract multi-scale information through repeated feature partitioning, iterative similarity computation and progressive feedback, the DEG module employs parallel multi-scale dilated convolutions (dilation=1, 2, 3) to captures multi-scale features directly. This design avoids redundant feature propagation, significantly reduces computational complexity and achieves superior efficiency in multi-scale feature extraction and integration. The proposed structure highlights both simplicity and effectiveness, thereby maintaining a low parameter count.}}

{\remark{In the DEC module, to further enhance the depth and flexibility of multi-scale feature extraction, the module employs a grouping mechanism to decompose the input channels. It utilizes multi-scale dilated convolution paths combined with \( 1 \times 1 \) convolutions for channel fusion, enabling more refined and integrated feature representations. This design enriches the diversity of feature expression, allowing for a more comprehensive perception of both local and global spatial information.}}

\subsection{Global Extension Embedding Branch}
The GEE branch adopts a global context aggregation strategy compared with the PEE branch, which progressively generates and enhances local embedded details. In the GEE branch, a mean pooling operation replaces the DEG module, processing the output features of the backbone \( \mathbf{f}_i \) to obtain a global feature representation. The global feature representation is then propagated to the DEC module for further refinement. 

By applying the mean pooling to \( \mathbf{f}_i \), the GEE branch provides a more global perspective of feature representation before diverse embedded local detail processing. The backbone output feature \( \mathbf{f}_i \in \mathbb{R}^{C_i \times H_i \times W_i} \) undergoes mean pooling to compute the global average feature representation as follows:
\begin{equation}
\mathbf{f}_{mi}^{+} = \frac{1}{H_i \times W_i} \sum_{h=1}^{H_i} \sum_{w=1}^{W_i} \mathbf{f}_i(h, w)
\label{eq5}
\end{equation}
where \(h=1,\cdots,H_i\) and \(w=1,\cdots,W_i\) are height factor and width factor, respectively. This operation compresses the spatial dimensions, where \(\mathbf{f}_{mi}^{+}\in \mathbb{R}^{C} \). The global feature is then fed into the DEC module for further processing to obtain \(\mathbf{f}_{mi}^{++}\) described as the DEC module \ref{DEC module}. In this way, the GEE branch can aggregate global information, generating comprehensive global feature embeddings that provide the DEC module with context-aware global features. Based on this global perspective, global and local consistency is ensured in the GEE branch.  
\subsection{ Cross-Domain Enhanced Alignment branch}
{\color{black}In CVGL tasks, images captured from different platforms often exhibit significant geometric differences, resulting in substantial variations in feature distributions and posing challenges for cross-domain matching. Feature alignment has been proven to be effective in addressing this issue \cite{shi2022accurate,tian2022smdt,zhang2024aligning}. Inspired by \cite{xia2024enhancing}, we designed a cross-domain enhanced alignment (CEA) branch that uses multi-level fusion and adaptive calibration strategies to dynamically adjust feature consistency within a shared latent space in high-dimensional embeddings. As shown in Fig. \ref{fig2}, after the initial feature \( \mathbf{f}_i\) extraction by the backbone, a transformation \( \mathcal{M}_{\text{i}}(\cdot) \) is applied to flatten the spatial dimensions \( H_i \times W_i \) into a single dimension \( L_i \) to obtain reshaped features \( \mathbf{f}_i^{\text{2D}} \in \mathbb{R}^{C_i \times L_i} \), which serve as the input features for the CEA branch. Here, \( L_i = H_i \times W_i \), \( i  \in\{d,s\}\). 

To enhance the consistency of cross-domain features, we introduce an adaptive calibration strategy in the CEA branch. Specifically, a \(1 \times 1\) convolutional layer \( \theta_q^1 \) is employed to project the two-dimensional input features \( \mathbf{f}^{\text{2D}}_i\) into a higher-dimensional space. Subsequently, the batch normalization operation and the ReLU activation function are used to adjust the feature distribution. This procedure obtains a high-dimensional feature representation \( \mathbf{f}_i^{\mathit{h}} \in \mathbb{R}^{2C_i \times L_i} \). Then, another \(1 \times 1\) convolutional layer \( \theta_v^1 \) is used to map the high-dimensional features back to the low-dimensional space, followed by the dropout operation and the normalization operation \(\mathcal{N}(\cdot)\) to improve the generalization capability. This produces a more compact feature representation \( \mathbf{f}_i^{\mathit{l}} \in \mathbb{R}^{C^{\prime} \times L_i} \), which improves computational efficiency and can be expressed as follows:
\setcounter{equation}{5}
\begin{equation}
\mathbf{f}_i^{\mathit{h}}  = \mathbf{F}_{ReLu}( \mathcal{B}( \theta_q^1( \mathbf{f}^{\text{2D}}_i )))
\end{equation}
\begin{equation}
\mathbf{f}_i^{\mathit{l}} = \mathcal{N} (\mathcal{D}( \theta_v^1 ( \mathbf{f}_i^{\mathit{h}} )))\
\end{equation}

After obtaining compact low-dimensional features \(\mathbf{f}_i^{\mathit{l}}\), a 1×1 convolutional layer \( \theta_k^1 \) is applied to further adjust the feature representation.  We apply T-Softmax \( \mathbf{F}_{\text{softmax}}^t(\cdot) \) to achieve the temperature-scaled local features with  temperature factor  \( T \), highlighting the crucial distinctions of the features   \(\mathbf{f}_i^{\mathit{l}}/ T\). This is followed by S-Softmax \( \mathbf{F}_{\text{softmax}}^s(\cdot) \) for global re-balancing to ensure that the features  \( \mathbf{F}_{\text{softmax}}^t ( \mathbf{f}_i^{\mathit{l}} / T )\)  can maintain consistency and harmony across different levels or scales. The output generated by the adaptive calibration strategy is denoted as \( \mathbf{f}_i^{\mathit{w}} \in \mathbb{R}^{C^{\prime} \times L_i} \) , which captures both local distinctions and global consistency. 

To integrate the original and calibrated features, multi-level fusion is employed, which enables the aggregation of complementary representations and enhances cross-domain feature consistency. Specifically, both the input feature \( \mathbf{f}^{\text{2D}}_i\) and the features  \( \mathbf{f}_i^{\mathit{w}}  \) are transformed into a shared dimension of \( \frac{C_i + C^{\prime}}{2} \) by  the \( 1 \times 1 \) convolutional layers \( \omega_q^1 \) and  \( \omega_v^1 \), respectively. Once the features have been transformed into the same dimensional space, they are concatenated along the channel dimension, namely the concatenated features represented as follows:
\begin{equation}
\mathbf{f}_i^c = \text{cat} (  \omega_q^1(\mathbf{f}_i), \omega_v^1(\mathbf{f}_i^{\mathit{w}}))
\end{equation}
where \( \mathbf{f}_i^{\mathit{c}} \in \mathbb{R}^{(C_i + C^{\prime}) \times L_i} \). Then a \( 1 \times 1 \) convolutional layer \( \omega_k^1 \) is applied to the concatenated features to restore the channel size back to \( \frac{C_i + C^{\prime}}{2} \), achieving a more compact representation. Finally, the batch normalization  and the ReLU activation function are employed to normalize and activate the feature distributions. The output of the  CEA branch is computed as follows:
\begin{equation}
\mathbf{f}_i^o = \mathbf{F}_{\text{ReLU}}( \mathcal{B}( \omega_k^1(\mathbf{f}_i^c)))
\label{eq10}
\end{equation}}

{\remark{The proposed CEA branch maps the input features to a higher-dimensional space to learn rich deep semantic information and optimize the feature distribution. Subsequently, a dual adaptive temperature scaling mechanism reconstructs the high-dimensional features, enhancing the saliency of key features and maintaining global consistency. On this basis, the branch fuses high- and low-dimensional features. Through two stages of dimensionality reduction, it compresses the feature representation and ultimately generates compact and highly expressive cross-domain aligned features.}}

\subsection{Multi-Loss Optimization}
In order to guide MEAN for learning, each of the three branches is optimized using different losses. The following will introduce the specific losses employed, including multi-level constrain named cross-domain invariant mapping alignment loss (CDA loss), InfoNCE loss \cite{oord2018representation}, and cross-entropy loss (CE loss).

\begin{figure}
  \centering
  \includegraphics[width=3.5in]{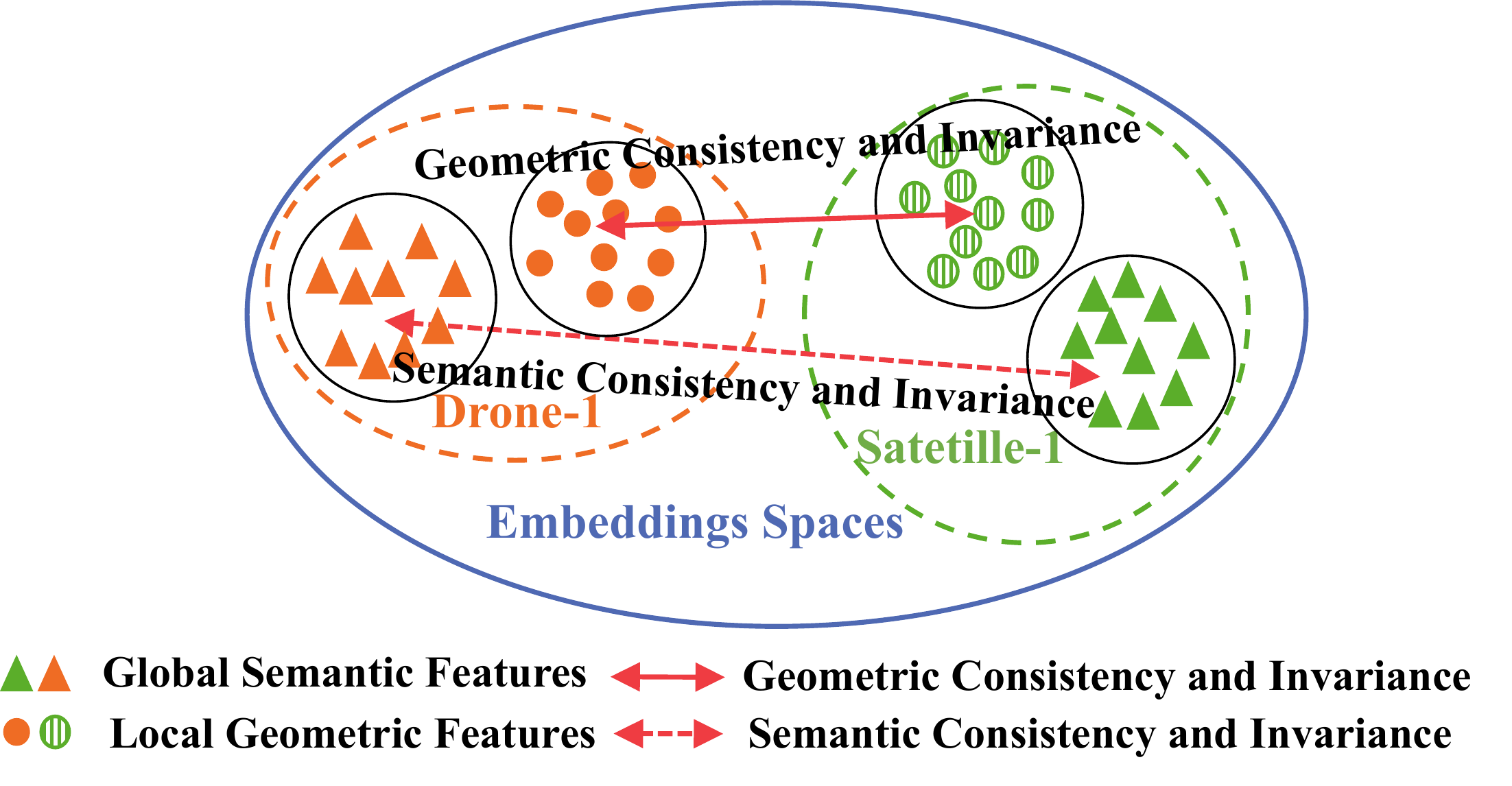}
  \caption{Illustration of Global Semantic and Local Geometric Feature Alignment Optimized by Our Proposed CDA Loss using Cosine Similarity and Mean Squared Error.}
  \label{fig3}  
  \end{figure}

\textit{CDA Loss}: The CEA branch effectively aligns features from different viewpoints through multi-level fusion and adaptive transformation. However, ensuring robust cross-view feature representation during consistency mining remains a significant challenge. To address this, we designed the CDA loss, as shown in Fig.\ref{fig3}, to learn multi-level features across viewpoints and mine feature alignment consistency and cross-domain invariance within a shared space.

For the feature embeddings \( \mathbf{f}_d^o \) and \( \mathbf{f}_s^o \) generated by the CEA branch described in Eq.(\ref{eq10}), cosine similarity \( \mathcal{C}(\cdot,\cdot) \) is used to estimate spatial directional consistency, while mean square errors measure absolute differences in the components of the feature.  Using cosine similarity and mean square errors, CDA loss effectively realizes the semantic consistency of aligned features at a global scale and maintains consistency in local details. The loss function can be formulated as follows:
\begin{equation}
\mathcal{L}_{\text{CDA}} = \alpha\mathcal{C}(\mathbf{f}_d^o, \mathbf{f}_s^o) +  \beta \mathcal{D}(\mathbf{f}_d^o, \mathbf{f}_s^o)
\end{equation}
where  \( \alpha \) and \( \beta \) are weighting coefficients to balance the contributions of cosine similarity  \( \mathcal{C}(\mathbf{f}_d^o, \mathbf{f}_s^o) \) and mean square errors \( \mathcal{D}(\mathbf{f}_d^o, \mathbf{f}_s^o) \) . \( \mathcal{C}(\mathbf{f}_d^o, \mathbf{f}_s^o) \)  is designed to promote global semantic consistency and invariance across cross-view features and represented as follows:
\begin{equation}
\mathcal{C}(\mathbf{f}_d^o, \mathbf{f}_s^o) = 1 - \frac{1}{M} \sum_{k=1}^{M} \frac{\mathbf{f}_{d_k}^o \cdot \mathbf{f}_{s_k}^o}{\|\mathbf{f}_{d_k}^o\| \|\mathbf{f}_{s_k}^o\|}
\end{equation}
where \(d_k\) and \(s_k\), \(k=1,\cdots,M\) are denoted index of the aerial and satellite images, respectively. \(M\) is the number of images we have used for each view in the training phase. \( \mathcal{D}(\mathbf{f}_d^o, \mathbf{f}_s^o) \) is  used to promote local semantic consistency and invariance between \(\mathbf{f}_d^o\) and \(\mathbf{f}_s^o \), and  defined as follows:
\begin{equation}
\mathcal{D}(\mathbf{f}_d^o, \mathbf{f}_s^o) = 1 -\frac{1}{M} \sum_{k=1}^{M} \|\mathbf{f}_{d_k}^o - \mathbf{f}_{s_k}^o\|^2
\end{equation}

Therefore, CDA loss can obtain dual capacities  of the global feature alignment and the local detail alignment.

{\remark{CDA Loss is designed to enforce feature consistency from both global and local perspectives. The global consistency constraint ensures that the overall representations across domains remain aligned, while the local consistency constraint focuses on fine-grained feature alignment. This approach effectively captures invariant features under style transformations, thereby significantly enhancing the model's robustness.}}

\textit{InfoNCE Loss}: In the PEE branch, although diversity and hierarchical embedding generation have been achieved, there exists a limitation of the consistency and discriminative capability. To address this issue, we introduce the InfoNCE loss to optimize the embedding space. By contrasting embeddings from different scenes (negative samples) and embeddings from the same scene (positive samples), InfoNCE loss ensures the consistency of embedding features within the same scene while maximizing the discriminability between embeddings of different scenes. The InfoNCE loss contains two queries from both detections, one of which uses a known satellite image to query \(D\) aerial images, namely satellite \(\rightarrow \) drone. The other uses a known aerial image to query \(S\) satellite images, which can be termed as drone\(\rightarrow \)satellite. The InfoNCE Loss is defined as follows:
\begin{equation}
\begin{aligned}
\mathcal{L}_{\text{InfoNCE}}\!=\!-\! \frac{1}{2N} \sum& \!\Bigg(\!  \log \frac{\exp ( \mathbf{f}_{gs}^{++} \cdot \mathbf{f}_{gd_+}^{++} / \tau )}{\sum_{j=1}^{D} \exp ( \mathbf{f}_{gs}^{++} \cdot \mathbf{f}_{gdj}^{++} / \tau)} \\
 +& \log \frac{\exp( \mathbf{f}_{gd}^{++} \cdot \mathbf{f}_{gs_+}^{++} / \tau )}{\sum_{p=1}^{S} \exp ( \mathbf{f}_{gd}^{++} \cdot \mathbf{f}_{gsp}^{++} / \tau )} \Bigg)
\end{aligned}
\end{equation}
where \( \mathbf{f}_{gs}^{++}\) described in Eq.(\ref{eq04}) represents an encoded satellite image, which is referred to as the query.  \( D \) is the number of encoded aerial images \( \mathbf{f}_{gdj}^{++} \), \(j=1,\cdots,D\), called references, among which only one positive sample denoted as \( \mathbf{f}_{gd_+}^{++} \) matches the query \( \mathbf{f}_{gs}^{++}\). Similarly, \( \mathbf{f}_{gd}^{++}\) described in Eq.(\ref{eq04}) represents an encoded satellite image, which is also called the query.  \( S \) is the number of encoded satellite images \( \mathbf{f}_{gsp}^{++} \), \(p=1,\cdots,S\), among which only one positive sample denoted \( \mathbf{f}_{gs_+}^{++} \) matches the query \( \mathbf{f}_{gd}^{++}\). The operation \(\sum\) represents the sum of the loss for \( N \) samples during the training phase. $\tau$ is the temperature scaling parameter.

\textit{CE Loss}: In the GEE branch, the CE loss is introduced to optimize the global feature embeddings \( \mathbf{f}_{mi}^{++} \) described in Eq.(\ref{eq5}). The primary purpose of this loss function is to ensure that the model not only extracts rich semantic information from the global context, but also accurately maps these global feature embeddings \( \mathbf{f}_{mi}^{++} \) to the correct class labels, thereby improving the classification performance. Through CE loss, the model learns the correspondence between global features and class labels, minimizing the discrepancy between the predicted and true distributions. CE loss measures the difference between the predicted class distribution of the global feature embeddings \( \mathbf{f}_{mi}^{++} \) and the true class label. The loss function is defined as follows:
\begin{equation}
\mathcal{L}_{CE} = -\sum_{r=1}^{N} \log(\hat{p}(y_r | \mathbf{f}_{mi}^{++}))
\end{equation}
\begin{equation}
\hat{p}(y_r | \mathbf{f}_{mi}^{++}) = \frac{\exp(z_r(y_r))}{\sum_{c=1}^{C} \exp(z_r(c))}
\end{equation}
where \( z_r(y) \) represents the logit score corresponding to the class \( y_r \) for the feature embedding \( \mathbf{f}_{mi}^{++} \), and \( \hat{p}(y_r | \mathbf{f}_{mi}^{++}) \) is the probability of the feature embedding belonging to the label \( y_r \). $C$ represents the total number of classes (i.e., the number of categories in the classification task), and $c$ denotes the index of all possible classes.

\textit{Total Loss}: In the overall model, we combine three loss functions \(\mathcal{L}_{\text{CDA}}\), \(\mathcal{L}_{\text{InfoNCE}}\), and \(\mathcal{L}_{\text{CE}}\). We optimize the overall performance of the model by minimizing these three functions. The total loss is formulated as follows:

\begin{equation}
\mathcal{L}_{total} = \lambda_1 \mathcal{L}_{\text{CDA}} + \lambda_2 \mathcal{L}_{\text{InfoNCE}} + \lambda_3 \mathcal{L}_{\text{CE}}
\end{equation}
where \(\lambda_1\), \(\lambda_2\) and \(\lambda_3\) are utilized to weigh the relative importance of the different loss terms.

\section{Experimental  results}\label{results}
\subsection{Experimental Datasets and Evaluation Metrics}
To evaluate the performance of our proposed CVGL framework, we conduct experiments on two large-scale datasets and one multi-weather scenario, including University-1652 \cite{zheng2020university}, SUES-200 \cite{zhu2023sues}, and Multi-weather University-1652 \cite{wang2024multiple}. These datasets provide complementary challenges for multi-view image matching and retrieval tasks.

\textit{University-1652} is a CVGL dataset consisting of drone, satellite, and ground-level images from 1,652 locations across 72 universities worldwide. The training set includes 701 buildings from 33 universities, while the test set consists of 951 buildings from 39 universities, with no overlap between training and testing data. This dataset is first used to introduce drone view to CVGL, extended the
visual localization task as ground-drone-satellite cross-view matching.

\textit{SUES-200} introduces altitude variation in aerial images, featuring 200 unique locations. The training and test sets contain 120 and 80 locations, respectively. Each location includes a satellite image and aerial images captured at four altitudes, encompassing diverse environments such as parks, lakes, and buildings. This dataset evaluates the model’s capability for cross-view retrieval with altitude variations.

\textit{Multi-weather University-1652} is an extension of University-1652 with ten simulated weather conditions providing a benchmark for evaluating model robustness under diverse environmental scenarios.

Furthermore, we employ Recall@K (R@K) and Average Precision (AP) as evaluation metrics. R@K measures the proportion of correct matches within the Top-\(K\)  retrieved results, while AP represents the balance between precision and recall. Additionally, we evaluate model efficiency through parameter count and computational complexity (GFLOPs) to reflect model portability under resource-constrained conditions. In the evaluation of parameter counts and model complexity, comparisons are conducted using the model states that achieve the optimal performance for each method.

{\remark{These datasets present unique challenges such as altitude variation (SUES-200), large-scale distractor sets (University-1652), and weather-induced feature distortion (Multi-weather University-1652)  to facilitate a comprehensive assessment of CVGL performance.}}
\begin{table*}[t]
  \centering
  \caption{Comparisons between the proposed method and some state-of-the-art methods on the University-1652 datasets. The best results are highlighted in red, while the second-best results are highlighted in blue.}
  \label{tab:comparison}
  \tiny
  \resizebox{\textwidth}{!}{
  \begin{tabular}{cccccccc}
  \hline
   &  &   &  & \multicolumn{2}{c}{Drone$\rightarrow$Satellite}  & \multicolumn{2}{c}{Satellite$\rightarrow$Drone}  \\ \cline{5-8} 
  \multirow{-2}{*}{Model} & \multirow{-2}{*}{{Venue}} & \multirow{-2}{*}{Parameters (M)} & \multirow{-2}{*}{GFLOPs} & {R@1} & {AP} & {R@1} & {AP} \\ \hline
  MuSe-Net\cite{wang2024multiple}      & PR’2024      & 50.47       &-          & 74.48       & 77.83      & 88.02     & 75.10 \\
  LPN\cite{wang2021each}           & TCSVT’2021   & 62.39       &\textcolor{blue}{36.78}     & 75.93       & 79.14      & 86.45     & 74.49 \\
  F3-Net\cite{sun2023f3}        & TGRS’2023    & -           &-         & 78.64       & 81.60      & -         & -     \\
  TransFG\cite{zhao2024transfg}       & TGRS’2024    &\textgreater{} 86.00       &-         & 84.01       & 86.31      & 90.16     & 84.61 \\
  IFSs\cite{ge2024multibranch}          & TGRS’2024    & -           &-         & 86.06       & 88.08      & 91.44     & 85.73 \\

  MCCG\cite{shen2023mccg}          & TCSVT’2023   & 56.65       &51.04     & 89.40       & 91.07      & 95.01     & 89.93 \\
  SDPL\cite{chen2024sdpl}          & TCSVT’2024   &\textcolor{blue}{42.56}       &69.71     & 90.16       & 91.64      & 93.58     & 89.45 \\
  MFJR\cite{ge2024multi}          & TGRS’2024    &\textgreater{} 88.00       &-         & 91.87       & 93.15      & 95.29     & 91.51 \\
  CCR\cite{du2024ccr}           & TCSVT2024     & 156.57      &160.61    & 92.54       & 93.78      & 95.15     & 91.80 \\
  Sample4Geo\cite{deuser2023sample4geo}    & ICCV’2023    & 87.57       &90.24     & 92.65       &93.81       & 95.14     & 91.39 \\
  SRLN\cite{lv2024direction}          & TGRS’2024    & 193.03      &-         &{92.70}   &{93.77}    & 95.14     & 91.97 \\
  DAC\cite{xia2024enhancing}           & TCSVT2024     & 96.50       & 90.24    &\textcolor{red}{94.67} &\textcolor{red}{95.50} &\textcolor{red}{96.43}&\textcolor{red}{93.79} \\ \hline
 MEAN(Ours)    & -            &\textcolor{red}{36.50}  & \textcolor{red}{26.18}   &\textcolor{blue}{93.55}       &\textcolor{blue}{94.53} &\textcolor{blue}{96.01}     & \textcolor{blue}{92.08}\\
  \hline
  \end{tabular}}
  \end{table*}

\subsection{Implementation Details}
We adopt a symmetric sampling strategy to select the input images. The \textit{ConvNeXt-Tiny} model, pre-trained on \textit{ImageNet}, is used as the backbone network for feature extraction with a newly added classifier module initialized by the Kaiming initialization method. During both training and testing, all input images are uniformly resized to \(3\times 384\times 384\). We also apply a series of data augmentation techniques, including random cropping, random horizontal flipping and random rotation. The batch size is set to 64 images (32 aerial images and 32 satellite images per batch). For optimization, we use the AdamW optimizer with an initial learning rate of 0.001. In \(\mathcal{L}_{\text{InfoNCE}}\), we employ \textit{label smoothing} with a smoothing factor of 0.1, and the temperature parameter \( \tau \) is set as a learnable parameter. Furthermore, we introduce two balancing factors \( \alpha \) and \( \beta\) to adjust the model's learning efficacy in \(\mathcal{L}_{\text{CDA}}\). All experiments are conducted on the Pytorch deep learning framework with the experimental platform running on Ubuntu 22.04 equipped with four NVIDIA RTX 4090 GPUs.

\subsection{Comparison with State-of-the-art Methods}
 We compare MEAN with state-of-the-art methods to demonstrate its effectiveness. The experimental results on the University-1652 dataset are reported in Table \ref{tab:comparison}, while the results on the Multi-weather University-165 dataset in Table \ref{tab:Multi-weather University-1652}. Additionally, the results pertaining to the SUES-200 dataset are shown in Tables \ref{tab:comparison SUES-200-1} and \ref{tab:comparison SUES-200-2}.
 
\textit{Results on University-1652}: As shown in Table \ref{tab:comparison}, MAEN achieves 93.55\% R@1 and 94.53\% AP in the Drone$\rightarrow$Satellite setting, and 96.01\% R@1 and 92.08\% AP in the Satellite$\rightarrow$Drone setting. These results demonstrate the strong performance and generalization capability of MAEN in CVGL. MAEN offers an efficient and lightweight solution, reducing model size by 62.17\% and computational complexity by 70. 99\% compared with the current state-of-the-art model DAC \cite{xia2024enhancing}, while maintaining comparable performance. Furthermore, with minimal parameter count and computational complexity, MAEN outperforms other advanced methods, including MCCG \cite{shen2023mccg} for multi-feature representation, MFJR \cite{ge2024multi} for multi-branch joint optimization, and Sample4Geo \cite{deuser2023sample4geo}, which employed the contrastive optimization techniques. These findings highlight MAEN's ability to achieve high accuracy while maintaining computational efficiency, making it an effective model for CVGL.
\begin{table*}[t]
\centering
  \caption{Comparison with state-of-the-art results under multi-weather conditions on the University-1652 dataset. The best results are highlighted in red, while the second-best results are highlighted in blue.}
\Large
  \label{tab:Multi-weather University-1652}
  \resizebox{\textwidth}{!}{
\begin{tabular}{ccccccccccc}

\hline
\multicolumn{1}{l}{} & Normal & Fog & Rain & Snow & Fog+Rain & Fog+Snow & Rain+Snow & Dark & \begin{tabular}[c]{@{}c@{}}Over\\ -exposure\end{tabular} & Wind \\ \cline{2-11} 
\multicolumn{1}{c}{\multirow{-2}{*}{Model}} & R@1/AP & R@1/AP & R@1/AP & R@1/AP & R@1/AP & R@1/AP & R@1/AP & R@1/AP & R@1/AP & R@1/AP \\ \hline
\multicolumn{11}{c}{Drone$\rightarrow$Satellite} \\ \hline
LPN\cite{wang2021each} & 74.33/77.60 & 69.31/72.95 & 67.96/71.72 & 64.90/68.85 & 64.51/68.52 & 54.16/58.73 & 65.38/69.29 & 53.68/58.10 & 60.90/65.27 & 66.46/70.35 \\
MuSeNet\cite{wang2024multiple} & 74.48/77.83 & 69.47/73.24 & 70.55/74.14 & 65.72/69.70 & 65.59/69.64 & 54.69/59.24 & 65.64/70.54 & 53.85/58.49 & 61.65/65.51 & 69.45/73.22 \\
Sample4Geo\cite{deuser2023sample4geo} & \textcolor{blue}{90.55/92.18} & \textcolor{blue}{89.72/91.48} & \textcolor{blue}{85.89/88.11} & \textcolor{blue}{86.64/88.18} & \textcolor{blue}{85.88/88.16} & \textcolor{blue}{84.64/87.11} & \textcolor{blue}{85.98/88.16} & \textcolor{blue}{87.90/89.87} & \textcolor{blue}{76.72/80.18} & \textcolor{blue}{83.39/89.51} \\
MEAN(Ours) & \textcolor{red}{90.81/92.32} & \textcolor{red}{90.97/92.52} & \textcolor{red}{88.19/90.05} & \textcolor{red}{88.69/90.49} & \textcolor{red}{86.75/88.84} & \textcolor{red}{86.00/88.22} & \textcolor{red}{87.21/89.21} & \textcolor{red}{87.90/89.87} & \textcolor{red}{80.54/83.53} & \textcolor{red}{89.27/91.01} \\ \hline
\multicolumn{11}{c}{Satellite$\rightarrow$Drone} \\ \hline
LPN\cite{wang2021each} & 87.02/75.19 & 86.16/71.34 & 83.88/69.49 & 82.88/65.39 & 84.59/66.28 & 79.60/55.19 & 84.17/66.26 & 82.88/52.05 & 81.03/62.24 & 84.14/67.35 \\
MuSeNet\cite{wang2024multiple} & 88.02/75.10 & 87.87/69.85 & 87.73/71.12 & 83.74/66.52 & 85.02/67.78 & 80.88/54.26 & 84.88/67.75 & 80.74/53.01 & 81.60/62.09 & 86.31/70.03 \\
Sample4Geo\cite{deuser2023sample4geo} & \textcolor{blue}{95.86/89.86} & \textcolor{blue}{95.72/88.95} & \textcolor{blue}{94.44/85.71} & \textcolor{red}{95.01}/\textcolor{blue}{86.73} & \textcolor{blue}{93.44/85.27} & \textcolor{blue}{93.72/84.78} & \textcolor{blue}{93.15/85.50} & \textcolor{blue}{96.01/87.06} & \textcolor{blue}{89.87/74.52} &\textcolor{blue}{95.29}/\textcolor{red}{87.06} \\
MEAN(Ours) & \textcolor{red}{96.58/89.93} & \textcolor{red}{96.00/89.49} & \textcolor{red}{95.15/88.87} & \textcolor{blue}{94.44}/\textcolor{red}{87.44} & \textcolor{red}{93.58/86.91} & \textcolor{red}{94.44/87.44} & \textcolor{red}{93.72/86.91} & \textcolor{red}{96.29/89.87} & \textcolor{red}{92.87/79.66} & \textcolor{red}{95.44/\textcolor{blue}{86.05}} \\ \hline
\end{tabular}}
\end{table*}

\textit{Results on Multi-weather University-1652}: As shown in Table \ref{tab:Multi-weather University-1652}, MEAN consistently achieves superior performance across various weather conditions in both Drone$\rightarrow$Satellite setting and Satellite$\rightarrow$Drone setting. In the Drone$\rightarrow$Satellite setting, MEAN achieves the best performance across all 10 weather conditions. It records R@1 and AP scores of 90.81\% and 92.32\% under Normal conditions, 90.97\% and 92.52\% under Fog, and 88.19\% and 90.05\% under Rain, significantly outperforming other methods. Under challenging conditions like Snow and Fog+Rain, MEAN continues to deliver strong results, achieving an R@1 of 88.69\% and AP of 90.49\% in Snow, and 86.75\% and 88.84\% in Fog+Rain.Results under other conditions show similar trends.
Similarly, in the Satellite$\rightarrow$Drone setting, MEAN sets new benchmarks in 8 out of 10 weather scenarios. It achieves R@1 scores of 96.58\%, 96.00\%, and 95.15\% under Normal, Fog, and Rain conditions, respectively. Even in adverse scenarios such as Over-exposure and Dark, MEAN maintains robust performance, reaching R@1 scores of 92.87\% and 96.29\%, respectively. Results under other conditions show similar trends. Compared with the state-of-the-art models such as Sample4Geo, LPN \cite{wang2021each}, and MuSeNet \cite{wang2024multiple}, MEAN demonstrates a remarkable balance between performance and robustness across diverse environmental conditions. These results highlight MEAN's ability to effectively learn feature consistency and maintain high accuracy, even under challenging cross-view and multi-weather scenarios.

\begin{table*}[t]
  \centering
  \small
  \caption{Comparisons between the proposed method and some state-of-the-art methods on the SUES-200 dataset in the Drone$\rightarrow$Satellite. The best results are highlighted in red, while the second-best results are highlighted in blue.}
  \label{tab:comparison SUES-200-1}
  \resizebox{\textwidth}{!}{
  \begin{tabular}{cccccccccccc}
  \hline
   &  &  &  & \multicolumn{8}{c}{Drone$\rightarrow$Satellite} \\ \cline{5-12} 
   &  &  &  & \multicolumn{2}{c}{150m} & \multicolumn{2}{c}{200m} & \multicolumn{2}{c}{250m} & \multicolumn{2}{c}{300m} \\ \cline{5-12} 
  \multirow{-3}{*}{Model} & \multirow{-3}{*}{Venue} & \multirow{-3}{*}{Parameters(M)} & \multirow{-3}{*}{GFLOPs} & \multicolumn{1}{c}{R@1} & \multicolumn{1}{c}{AP} & \multicolumn{1}{c}{R@1} & \multicolumn{1}{c}{AP} & \multicolumn{1}{c}{R@1} & \multicolumn{1}{c}{AP} & \multicolumn{1}{c}{R@1} & \multicolumn{1}{c}{AP} \\ \hline
  LPN\cite{wang2021each}                    & TCSVT’2022  & 62.39  &\textcolor{blue}{36.78}   & 61.58 & 67.23 & 70.85 & 75.96 & 80.38 & 83.80 & 81.47 & 84.53 \\
  IFSs\cite{ge2024multibranch}              & TGRS’2024   & -      &-       & 77.57 & 81.30 & 89.50 & 91.40 & 92.58 & 94.21 & 97.40 & 97.92 \\
  MCCG\cite{shen2023mccg}                   & TCSVT’2023  & 56.65  &51.04   & 82.22 & 85.47 & 89.38 & 91.41 & 93.82 & 95.04 & 95.07 & 96.20 \\
  SDPL\cite{chen2024sdpl}                   & TCSVT’2024  & \textcolor{blue}{42.56}  &69.71   & 82.95 & 85.82 & 92.73 & 94.07 & 96.05 & 96.69 & 97.83 & 98.05 \\
  CCR\cite{du2024ccr}                       & TCSVT’2024  & 156.57 &160.61  & 87.08 & 89.55 & 93.57 & 94.90 & 95.42 & 96.28 & 96.82 & 97.39 \\
  MFJR\cite{ge2024multi}                    & TGRS’2024   & \textgreater{}88.00  &- & 88.95 & 91.05 & 93.60 & 94.72 & 95.42 & 96.28 & 97.45 & 97.84 \\
  SRLN\cite{lv2024direction}                & TGRS’2024   & 193.03 &-       & 89.90 & 91.90 & 94.32 & 95.65 & 95.92 & 96.79 & 96.37 & 97.21 \\
  Sample4Geo\cite{deuser2023sample4geo}     & ICCV’2023   & 87.57  &90.24   & 92.60 & 94.00 & 97.38 & 97.81 & \textcolor{blue}{98.28} & \textcolor{blue}{98.64} & \textcolor{blue}{99.18} & \textcolor{blue}{99.36} \\
  DAC\cite{xia2024enhancing}                & TCSVT’2024  & 96.50 &90.24  &\textcolor{red}{96.80}  &\textcolor{red}{97.54} & \textcolor{blue}{97.48} & \textcolor{blue}{97.97} & 98.20 & 98.62 & 97.58 & 98.14 \\\hline
  MEAN(Ours)      & -           &\textcolor{red}{36.50}  &\textcolor{red}{26.18}  &\textcolor{blue}{95.50} &\textcolor{blue}{96.46} &\textcolor{red}{98.38}&\textcolor{red}{98.72}& \textcolor{red}{98.95} &\textcolor{red}{99.17}  &\textcolor{red}{99.52}  & \textcolor{red}{99.63} \\ \hline
  \end{tabular}}
  \end{table*}
\begin{table*}[ht]
  \centering
  \small
  \caption{{\color{black}Comparisons between the proposed method and some state-of-the-art methods on the SUES-200 dataset in the Satellite$\rightarrow$Drone. The best results are highlighted in red, while the second-best results are highlighted in blue.}}
  \label{tab:comparison SUES-200-2}
  \resizebox{\textwidth}{!}{
  \begin{tabular}{cccccccccccc}
  \hline
   &  &  &  & \multicolumn{8}{c}{Satellite$\rightarrow$Drone} \\ \cline{5-12} 
   &  &  &  & \multicolumn{2}{c}{150m} & \multicolumn{2}{c}{200m} & \multicolumn{2}{c}{250m} & \multicolumn{2}{c}{300m} \\ \cline{5-12} 
  \multirow{-3}{*}{Model} & \multirow{-3}{*}{Venue} & \multirow{-3}{*}{Parameters(M)} & \multirow{-3}{*}{GFLOPs} & R@1 & AP & R@1 & AP & R@1 & AP & R@1 & AP \\ \hline
  LPN\cite{wang2021each}       & TCSVT’2022       & 62.39  &\textcolor{blue}{36.78} & 83.75 & 83.75 & 83.75 & 83.75 & 83.75 & 83.75 & 83.75 & 83.75 \\
  CCR\cite{du2024ccr}          & TCSVT’2024       & 156.57 &160.61 & 92.50 & 88.54 & 97.50 & 95.22 & 97.50 & 97.10 & 97.50 & 97.49 \\

  IFSs\cite{ge2024multibranch} & TGRS’2024        & -      &-  & 93.75 & 79.49 & 97.50 & 90.52 & 97.50 & 96.03 & \textcolor{red}{100.00} & 97.66 \\
  MCCG\cite{shen2023mccg}      & TCSVT’2023       & 56.65  &51.04 & 93.75 & 89.72 & 93.75 & 92.21 & 96.25 & 96.14 &\textcolor{blue}{ 98.75} & 96.64 \\
  SDPL\cite{chen2024sdpl}      & TCSVT’2024       & \textcolor{blue}{42.56}  &69.71 & 93.75 & 83.75 & 96.25 & 92.42 & 97.50 & 95.65 & 96.25 & 96.17 \\
  SRLN\cite{lv2024direction}   & TGRS’2024        & 193.03 &- & 93.75 & 93.01 & 97.50 & 95.08 & 97.50 & 96.52 & 97.50 & 96.71 \\
  MFJR\cite{ge2024multi}       & TGRS’2024        & \textgreater{}88.00 &-& 95.00 & 89.31 & 96.25 & 94.72 & 94.69 & 96.92 &\textcolor{blue}{98.75} & 97.14 \\
Sample4Geo\cite{deuser2023sample4geo} & ICCV’2023        & 87.57  &90.24 &\textcolor{red}{97.50} &93.63 &\textcolor{blue}{98.75} &\textcolor{blue}{96.70} &\textcolor{blue}{98.75} &\textcolor{blue}{98.28} &\textcolor{blue}{98.75} &\textcolor{blue}{98.05}\\
  DAC\cite{xia2024enhancing}   & TCSVT’2024       & 96.50 &90.24 &\textcolor{red}{97.50} &\textcolor{blue}{94.06}  & \textcolor{blue}{98.75} & 96.66 &\textcolor{blue}{98.75} & 98.09 &\textcolor{blue}{98.75} & 97.87 \\\hline
  MEAN(Ours) & -                & \textcolor{red}{36.50}  &\textcolor{red}{26.18} &\textcolor{red}{97.50}  & \textcolor{red}{94.75} &\textcolor{red}{100.00}  & \textcolor{red}{97.09} & \textcolor{red}{100.00} &\textcolor{red}{98.28}  & \textcolor{red}{100.00} &\textcolor{red}{ 99.21} \\ \hline
  \end{tabular}}
\end{table*}
\begin{table*}[t]
\centering
\small
\caption{Comparisons between the proposed method and state-of-the-art methods in cross-domain evaluation on Drone$\rightarrow$Satellite. The best results are highlighted in red, while the second-best results are highlighted in blue.}
\label{tab:comparison University-SUES-1}
\resizebox{\textwidth}{!}{
\begin{tabular}{cccccccccccc}
\hline
 &  &  &  & \multicolumn{8}{c}{Drone$\rightarrow$Satellite} \\ \cline{5-12} 
 &  &  &  & \multicolumn{2}{c}{150m} & \multicolumn{2}{c}{200m} & \multicolumn{2}{c}{250m} & \multicolumn{2}{c}{300m} \\ \cline{5-12} 
\multirow{-3}{*}{Model} & \multirow{-3}{*}{Venue} & \multirow{-3}{*}{Parameters(M)} & \multirow{-3}{*}{GFLOPs} & R@1 & AP & R@1 & AP & R@1 & AP & R@1 & AP \\ \hline
MCCG\cite{shen2023mccg}      & TCSVT’2023 &\textcolor{blue}{56.65}  &\textcolor{blue}{51.04}& 57.62 & 62.80 & 66.83 & 71.60 & 74.25 & 78.35 & 82.55 & 85.27  \\
Sample4Geo\cite{deuser2023sample4geo}& ICCV’2023  & 87.57  &90.24& 70.05 & 74.93 & 80.68 & 83.90 & 87.35 & 89.72 & 90.03 & 91.91   \\
DAC\cite{xia2024enhancing}       & TCSVT’2024 & 96.50 &90.24& 76.65 & 80.56 & 86.45 & 89.00 &\textcolor{red}{92.95}  & \textcolor{red}{94.18} & \textcolor{blue}{94.53} & \textcolor{blue}{95.45}   \\\hline
MEAN(Ours)  & -          & \textcolor{red}{36.50}  &\textcolor{red}{26.18}& \textcolor{red}{81.73} &\textcolor{red}{87.72}  &\textcolor{red}{89.05}  &\textcolor{red}{91.00}  &\textcolor{blue}{92.13}  &\textcolor{blue}{93.60}  &\textcolor{red}{94.63}  &\textcolor{red}{95.76}    \\ \hline
\end{tabular}}
\end{table*}

\begin{table*}[t]
\centering
\small
\caption{Comparisons between the proposed method and state-of-the-art methods in cross-domain evaluation on Satellite$\rightarrow$Drone. The best results are highlighted in red, while the second-best results are highlighted in blue.}
\label{tab:comparison University-SUES-2}
\resizebox{\textwidth}{!}{
\begin{tabular}{cccccccccccc}
\hline
 &  &  &  & \multicolumn{8}{c}{Satellite$\rightarrow$Drone} \\ \cline{5-12} 
 &  &  &  & \multicolumn{2}{c}{150m} & \multicolumn{2}{c}{200m} & \multicolumn{2}{c}{250m} & \multicolumn{2}{c}{300m} \\ \cline{5-12} 
\multirow{-3}{*}{Model} & \multirow{-3}{*}{Venue} & \multirow{-3}{*}{Parameters(M)} & \multirow{-3}{*}{GFLOPs} & R@1 & AP & R@1 & AP & R@1 & AP & R@1 & AP \\ \hline
MCCG\cite{shen2023mccg}    & TCSVT’2023  &\textcolor{blue}{56.65}  &\textcolor{blue}{51.04}& 61.25 & 53.51 & 82.50 & 67.06 & 81.25 & 74.99 & 87.50 & 80.20   \\
Sample4Geo\cite{deuser2023sample4geo} & ICCV’2023  & 87.57  &90.24& 83.75 & 73.83 & 91.25 & 83.42 &\textcolor{blue}{ 93.75} & 89.07 & 93.75 & 90.66   \\
DAC\cite{xia2024enhancing}        & TCSVT’2024 & 96.50 &90.24& \textcolor{blue}{87.50} &\textcolor{blue}{ 79.87} &\textcolor{red}{96.25} &\textcolor{blue}{ 88.98}  &\textcolor{red}{ 95.00} & \textcolor{red}{ 92.81} &\textcolor{red}{96.25} & \textcolor{blue}{94.00}   \\\hline
MEAN(Ours) & -          & \textcolor{red}{36.50}   &\textcolor{red}{26.18}& \textcolor{red}{91.25}   & \textcolor{red}{81.50}   & \textcolor{red}{96.25}   &  \textcolor{red}{89.55}  &\textcolor{red}{95.00} &\textcolor{blue}{ 92.36} &\textcolor{red}{96.25}  & \textcolor{red}{94.32}   \\ \hline
\end{tabular}}
\end{table*}

\textit{Results on SUES-200}:
As shown in Table \ref{tab:comparison SUES-200-1}, in the Drone$\rightarrow$Satellite setting, MEAN achieves R@1 scores of 95.50\%, 98.38\%, 98.95\%, and 99.52\% and AP scores of 96.46\%, 98.72\%, 99.17\%, and 99.63\% at different altitude settings (150m, 200m, 250m, 300m). Although MEAN narrowly falls behind other models at 150m in terms of R@1 and AP, it surpasses state-of-the-art models at the other three altitude levels, achieving the best performance. This demonstrates MEAN’s strong adaptability in higher-altitude domains, effectively preserving high-level semantic consistency and robust across different views. Similarly, as shown in Table \ref{tab:comparison SUES-200-2}, in the Satellite$\rightarrow$Drone setting, MEAN achieves R@1 scores of 97.50\%, 100.00\%, 100.00\%, and 100.00\%, and AP scores of 94.75\%, 97.09\%, 98.28\%, and 99.21\% across these altitude levels. MEAN consistently achieves the best performance at all heights, further confirming its stability and robustness in extracting and matching cross-view image features, regardless of altitude variations. In general, MEAN demonstrates substantial advantages in both performance and efficiency. Compared with the state-of-the-art models such as Sample4Geo, DAC, MEAN achieves a significant reduction in parameter count and computational complexity while maintaining outstanding accuracy. Among 18 evaluation metrics, MEAN achieves the best results in 16 metrics and ranks second in the remaining two, highlighting its robust ability to learn feature consistency and invariance when addressing cross-view perspective differences and scale variations. In addition to its ability to learn consistent and invariant features between cross-view images, MEAN exhibits high computational efficiency, making it an effective solution for CVGL in large-scale scenarios.
\subsection{Comparison with State-of-the-Art Methods on Cross-Domain Generalization Performance}
In CVGL, cross-domain adaptability is a critical metric for evaluating the generalization capability of a model, especially when the training and testing datasets exhibit significant differences. To evaluate the transferability of our proposed model, we conducted experiments by training on the University-1652 dataset and testing on the SUES-200 dataset.

As shown in Table \ref{tab:comparison University-SUES-1} and Table \ref{tab:comparison University-SUES-2}, MEAN demonstrates exceptional cross-domain adaptability in both Drone$\rightarrow$Satellite setting and Satellite$\rightarrow$Drone setting across varying altitudes (150m, 200m, 250m, and 300m). In the Drone$\rightarrow$Satellite setting (Table \ref{tab:comparison University-SUES-1}), MEAN achieves competitive results at all altitudes. At 150m, it attains an R@1 of 81.73\% and an AP of 87.72\%, surpassing other methods. As altitude increases, MEAN maintains robust performance, achieving R@1/AP scores of 89.05\%/91.00\% at 200m, 92.13\%/93.60\% at 250m and 94.63\%/95.76\% at 300m. These results highlight MEAN's strong generalization capability across varying domains. In the Satellite$\rightarrow$Drone setting (Table \ref{tab:comparison University-SUES-2}), MEAN similarly exhibits superior generalization performance. At 150m, it achieves the highest R@1 of 91.25\% and AP of 81.50\%, maintaining high competitiveness at higher altitudes. At 200m, 250m and 300m, MEAN outperforms other models with R@1/AP scores of 96.25\%/89.55, 95.00\%/92.36\% and 96.25\%/94.32\%, respectively, further validating its robustness in CVGL. Compared with the state-of-the-art models such as Sample4Geo, DAC, MEAN achieves superior performance while significantly reducing model complexity. With only 36.50M parameters and 26.18 GFLOPs, MEAN achieves the best performance in 13 out of 18 evaluation metrics and ranks second in four metrics. These findings underscore MEAN's ability to learn domain-invariant features and adapt effectively to unseen data domains, making it a highly efficient and accurate model for CVGL under cross-domain conditions.
\vspace{-3pt}
\subsection{Ablation Studies}
In the ablation study, we evaluate the contribution of each component in MEAN on the University-1652 dataset. As shown in Table \ref{Ablation Studies}, the PEE branch, GEE branch, CEA branch, and CDA Loss are analyzed to assess their individual and combined impacts. Here, $\mathcal{L}(D)$ and $\mathcal{L}(C)$ represent the mean squared error and cosine similarity constraints within the CDA Loss, respectively. The baseline model is defined as ConvNeXt-Tiny with the CE loss trained solely without incorporating any additional branches.

The results demonstrate that incorporating the PEE branch individually improves performance. Combining PEE and GEE further enhances R@1 and AP, highlighting the complementary advantages of progressive diversification expansion and global extension embedding in improving feature representation. The addition of the CEA branch significantly enhances performance in both settings (Drone$\rightarrow$Satellite setting and Satellite$\rightarrow$Drone setting), indicating that the cross-domain enhanced alignment branch effectively aligns cross-domain features and strengthens domain invariance.
\begin{table}[ht]
  \Large
  \renewcommand{\arraystretch}{1.0}
  \caption{The influence of each component on the performance of proposed MEAN. The best results are highlighted in red.}
\resizebox{0.5\textwidth}{!}{
  \begin{tabular}{ccccccccc}
  \hline
  \multicolumn{5}{c}{\multirow{2}{*}{Setting}} & \multicolumn{4}{c}{University-1652}                                       \\ \cline{6-9} 
  \multicolumn{5}{c}{}                         & \multicolumn{2}{c}{Drone$\rightarrow$Satellite} & \multicolumn{2}{c}{Satellite$\rightarrow$Drone} \\ \hline
PEE      & GEE         & CEA     &\(\mathcal{L}\)(\(\mathcal{D}\))       &\(\mathcal{L}\)(\(\mathcal{C}\))                & R@1              & AP               & R@1              & AP               \\ \hline
           &             &         &       &                & 84.00            & 86.51            & 92.29            & 82.90            \\
\checkmark &             &         &       &                & 91.10            & 92.59            & 95.57            & 90.23            \\
\checkmark & \checkmark  &         &       &                & 91.49            & 92.93            & 95.72            & 90.80            \\
\checkmark &             & \checkmark      & \checkmark     &                  & 92.07            & 93.39            & 95.58            & 91.57            \\
\checkmark &             & \checkmark      & \checkmark     & \checkmark       & 92.84            & 94.04            & 95.44            & 91.78            \\
\checkmark & \checkmark  & \checkmark      & \checkmark     & \checkmark       &\textcolor{red}{93.55}            & \textcolor{red}{94.53}            & \textcolor{red}{96.01}            & \textcolor{red}{92.08}            \\ \hline
\label{Ablation Studies}
\end{tabular}}
\vspace{-3pt}
\end{table}

For loss functions, the joint application of $\mathcal{L}(D)$ and $\mathcal{L}(C)$ further improves the performance on cross-domain tasks. With all components and losses integrated, MEAN achieves the optimal results: R@1 and AP scores of 93.55\% and 94.53\% in the Drone$\rightarrow$Satellite setting, and 96.01\% and 92.08\% in the Satellite$\rightarrow$Drone setting. Compared with the baseline, MEAN improves R@1 and AP by 9.55\% and 7.92\% in the Drone$\rightarrow$Satellite setting, and by 3.79\% and 9.28\% in the Satellite$\rightarrow$Drone setting. These results validate the effectiveness of each component.
\begin{table}[t]
  \centering
  \caption{Performance impact of different dilation rates \(x\) in the DEG module with the DEC module fixed. The best results are highlighted in red, while the second-best results are highlighted in blue.}
\renewcommand{\arraystretch}{1.0}
  \resizebox{0.5\textwidth}{!}{
  \begin{tabular}{ccccccc}
  \hline  
  \multicolumn{3}{c}{\multirow{2}{*}{Setting (DEG)}} & \multicolumn{4}{c}{University-1652}                                       \\ \cline{4-7} 
  \multicolumn{3}{c}{}                         & \multicolumn{2}{c}{Drone$\rightarrow$Satellite} & \multicolumn{2}{c}{Satellite$\rightarrow$Drone} \\ \hline
\( \varphi^x_q \)     &\( \varphi^x_k \)         &\( \varphi^x_v \)   &R@1   &AP &R@1   &AP\\ \hline
  $x=1$  & $x=1$      & $x=1$  & 93.09            & 94.19            & 95.86            & \textcolor{blue}{92.72}            \\
  $x=2$  & $x=2$      & $x=2$  & 92.24            & 93.46            & 95.29            & 91.96            \\
  $x=3$  & $x=3$      & $x=3$  & 92.61            & 93.75            & 95.72            & 92.38            \\       
  $x=1$  & $x=1$      & $x=2$  & \textcolor{blue}{93.32}            & \textcolor{blue}{94.33}            & 95.44            & 92.51            \\
  $x=1$  & $x=1$      & $x=3$  & 93.14            & 94.16            & \textcolor{red}{96.14} & 92.40        \\
  $x=2$  & $x=2$      & $x=1$  & 92.41            & 93.61            & 96.00            & 91.98            \\
  $x=2$  & $x=2$      & $x=3$  & 93.16            & 94.23            & 95.86     & \textcolor{red}{92.76}    \\
  $x=3$  & $x=3$      & $x=1$  & 92.87            & 93.94            & 95.29            & 92.28            \\
  $x=3$  & $x=3$      & $x=2$  & 92.64            & 93.77            & 95.44            & 92.14            \\
  $x=1$  & $x=2$      & $x=3$  & \textcolor{red}{93.55}  &\textcolor{red}{94.54}  & \textcolor{blue}{96.01}  &{92.08}  \\ \hline
\label{Ablation Studies1}
\end{tabular}}
\end{table}
\begin{table}[t]
  \centering
  \caption{Performance impact of different dilation rates \(x\) in the DEC module with the DEG module fixed. The best results are highlighted in red, while the second-best results are highlighted in blue.}
\renewcommand{\arraystretch}{1.0}
\resizebox{0.5\textwidth}{!}{
  \begin{tabular}{ccccccc}
  \hline
  \multicolumn{3}{c}{\multirow{2}{*}{Setting (DEC)}} & \multicolumn{4}{c}{University-1652}                                       \\ \cline{4-7} 
  \multicolumn{3}{c}{}                         & \multicolumn{2}{c}{Drone$\rightarrow$Satellite} & \multicolumn{2}{c}{Satellite$\rightarrow$Drone} \\ \hline
\( \psi^x_q \)     &\( \psi^x_k \)         &\( \psi^x_v \)   &R@1   &AP &R@1   &AP\\ \hline
  $x=1$  & $x=1$      & $x=1$  & 93.24            & 94.30            & 95.43            & 92.55            \\
  $x=2$  & $x=2$      & $x=2$  & \textcolor{blue}{93.32}            &\textcolor{blue}{94.38}            & 95.72            & \textcolor{blue}{92.76}            \\
  $x=3$  & $x=3$      & $x=3$  & 93.29            & 94.36            & 95.15            & 92.62            \\
  $x=1$  & $x=1$      & $x=2$  & 92.92            & 94.06            & 95.29            & 92.56            \\
  $x=1$  & $x=1$      & $x=3$  & 93.19            & 94.31            & \textcolor{red}{96.15} & \textcolor{red}{92.80}        \\
  $x=2$  & $x=2$      & $x=1$  & 93.18            & 94.29            & 95.58            & 92.54            \\
  $x=2$  & $x=2$      & $x=3$  & 93.11            & 94.20            & 95.58     & 92.27    \\
  $x=3$  & $x=3$      & $x=1$  & 92.81            & 93.93            & 95.29            & 92.71            \\
  $x=3$  & $x=3$      & $x=2$  & 93.04            & 94.12            & 95.72            & \textcolor{blue}{92.76}            \\
  $x=1$  & $x=2$      & $x=3$  & \textcolor{red}{93.55}  &\textcolor{red}{94.54}  & \textcolor{blue}{96.01}  &{92.08}  \\ \hline
\label{Ablation Studies2}
\end{tabular}}
\end{table}

{\color{black}
To further investigate the effectiveness of using different dilation rates in the DEG and DEC modules, the dilation rates have been adjusted to various configurations. As shown in Table \ref{Ablation Studies1}, when the dilation rates of the DEC module are fixed at {1,2,3}, the experimental performance demonstrates that uniform configurations with same dilation rates in DEG module generally lead to suboptimal performance compared with the configurations with various dilation rates. From this table, the progressive pattern with dilation rates {1,2,3} in DEG module achieves the better results. Similarly, it can be observed that the progressive pattern with dilation rates {1,2,3} in DEC module also achieves the better results when the dilation rates of the DEG module are fixed at {1,2,3} in Table \ref{Ablation Studies2}. Furthermore, we conduct the experiment using the same dilation rates in both DEG and DEC modules as shown in Table \ref{Ablation Studies3}. The results demonstrate that when both modules share the varied dilation rate combinations, the model outperforms configurations with a single fixed dilation rate. Moreover, setting both modules to {1,2,3} achieves the best overall results.}
\begin{table}[ht]
  \centering
  \caption{Performance impact of   synchronously tuning the same dilation rates in the DEG and DEC modules. The best results are highlighted in red, while the second-best results are highlighted in blue.}
\renewcommand{\arraystretch}{1.0}
\resizebox{0.5\textwidth}{!}{
  \begin{tabular}{ccccccc}
  \hline
  \multicolumn{3}{c}{\multirow{2}{*}{Setting (DEG+DEC)}} & \multicolumn{4}{c}{University-1652}                                       \\ \cline{4-7} 
  \multicolumn{3}{c}{}                         & \multicolumn{2}{c}{Drone$\rightarrow$Satellite} & \multicolumn{2}{c}{Satellite$\rightarrow$Drone} \\ \hline
\(\varphi^x_q\psi^x_q \)     &\(\varphi^x_k\psi^x_k \)         &\(\varphi^x_v\psi^x_v \)   &R@1   &AP &R@1   &AP\\ \hline
  $x=1$  & $x=1$      & $x=1$  & 93.44            & 94.47            & 95.15            & 92.58            \\
  $x=2$  & $x=2$      & $x=2$  & 93.20            & 94.31            & 95.95            & \textcolor{red}{92.90}            \\
  $x=3$  & $x=3$      & $x=3$  & 92.60            & 93.80            & 95.86            & 91.96            \\
  $x=1$  & $x=1$      & $x=2$  & \textcolor{blue}{93.48}            & \textcolor{blue}{94.49}            & 95.72            & \textcolor{blue}{92.73}            \\
  $x=1$  & $x=1$      & $x=3$  & 93.12            & 94.20            & \textcolor{red}{96.14} & 92.65        \\
  $x=2$  & $x=2$      & $x=1$  & 93.26            & 94.30            & 95.72            & 92.52            \\
  $x=2$  & $x=2$      & $x=3$  & 93.14            & 94.22            & 95.01     & 92.21    \\
  $x=3$  & $x=3$      & $x=1$  & 92.84            & 93.99            & 95.29            & 92.28            \\
  $x=3$  & $x=3$      & $x=2$  & 93.29            & 94.38            & 95.29            & 92.32            \\
  $x=1$  & $x=2$      & $x=3$  & \textcolor{red}{93.55}  &\textcolor{red}{94.54}  & \textcolor{blue}{96.01}  &{92.08}  \\ \hline
\label{Ablation Studies3}
\end{tabular}}
\end{table}
\subsection{Feature Distribution}
\begin{figure*}[!ht]
    \centering
    \includegraphics[width=7.1in]{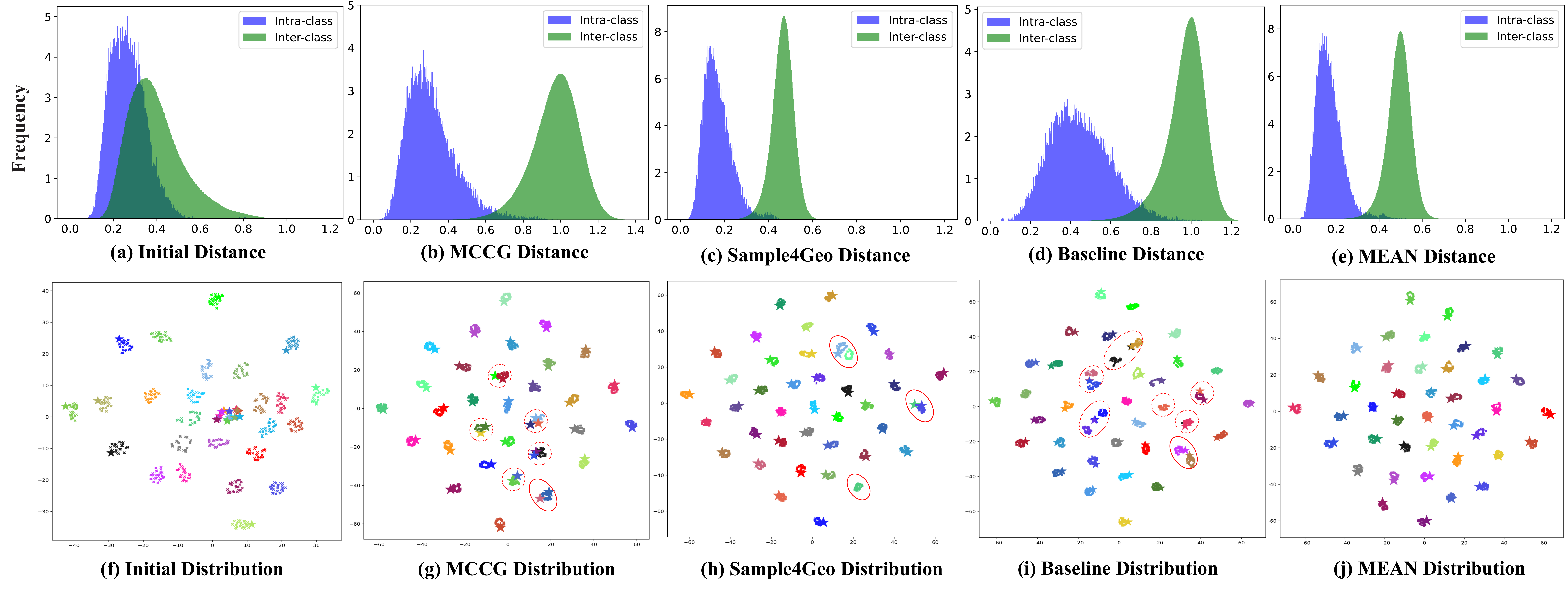}
    \caption{In the cross-view drone navigation task, (a-e) illustrate the intra-class and inter-class distances of features, where intra-class and inter-class distances are represented in blue and green, respectively. (f-j) depict the distribution of feature embeddings in the 2D feature space, with × and pentagrams representing aerial image features and satellite image features, respectively. A total of 40 locations were selected from the test set. Samples with the same color belong to the same location, while those with different colors indicate different locations.}
    \label{fig4} 
\end{figure*}
To comprehensively evaluate the effectiveness of the MEAN model in CVGL, we visualized the intra-class and inter-class distance distributions on the University-1652 dataset, as shown in Fig.~\ref{fig4} (a-e). The analysis includes a comparison of two representative methods, MCCG and Sample4Geo, along with a baseline model to validate the contributions of each component in MEAN. Compared with the initial features (Fig.~\ref{fig4}(a)), MCCG (Fig.~\ref{fig4}(b)), Sample4Geo (Fig.~\ref{fig4}(c)), and the baseline model (Fig.~\ref{fig4}(d)), the MEAN model achieves substantial separation between intra-class and inter-class distances, resulting in more compact intra-class features and more distinctly separated inter-class features. However, both the baseline model and the MCCG exhibit limitations in intra-class compactness, inter-class separation, and cross-view consistency. Although Sample4Geo demonstrates improved discriminative ability and consistency over the baseline and MCCG, it still exhibits overlap in some areas, indicating suboptimal intra-class compactness and inter-class separation. By contrast, the MEAN model (Fig.~\ref{fig4}(e)) significantly reduces intra-class distances while increasing inter-class distances, showcasing superior feature discriminability and cross-view consistency.
\begin{figure}[ht]
  \centering
  \includegraphics[width=3.4in]{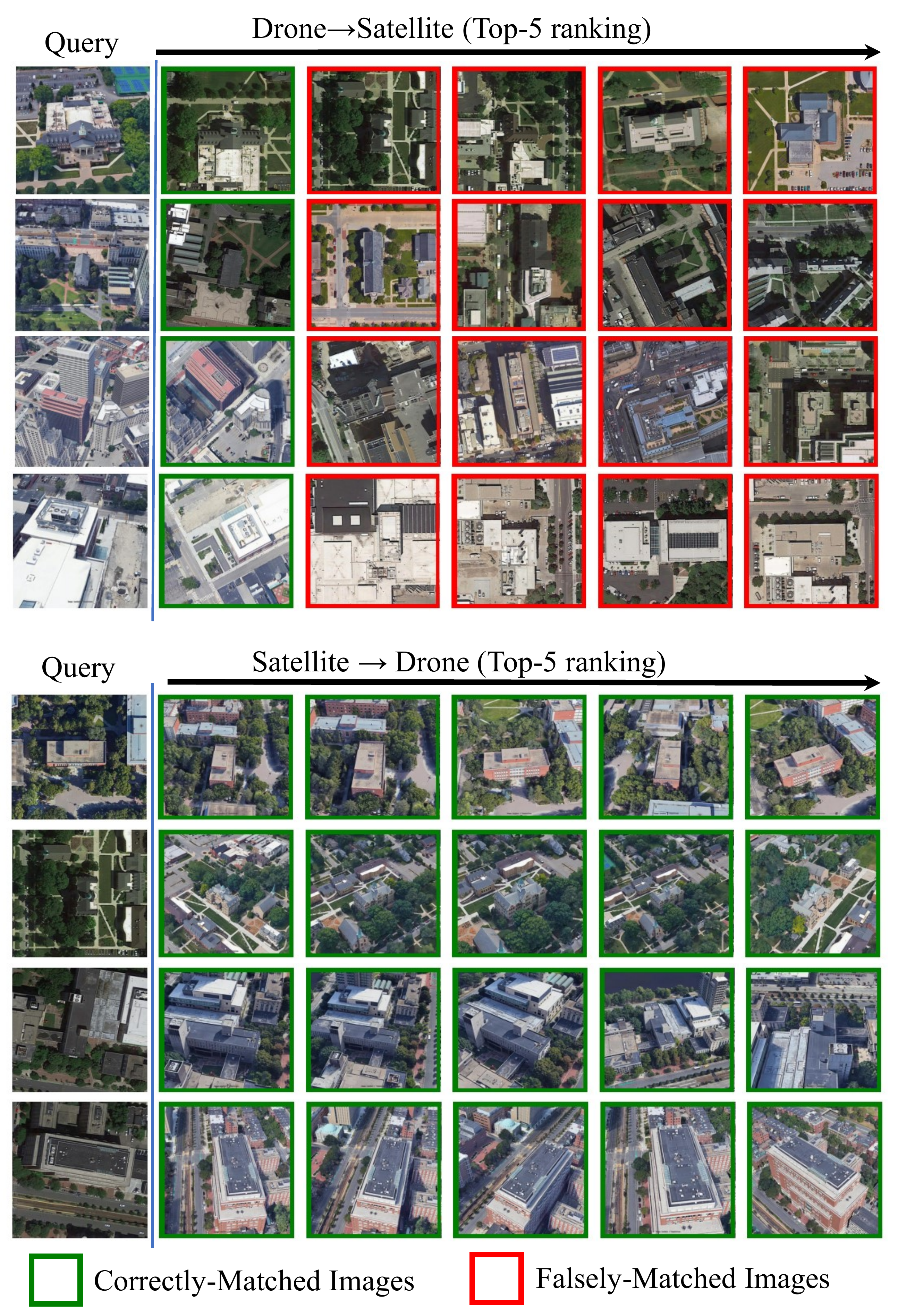}
  \caption{Top-5 Retrieval Results of the Proposed MEAN on the University-1652 Dataset.}
  \label{fig5}  
\end{figure}
\begin{figure}[ht]
    \centering
    \includegraphics[width=3.5in]{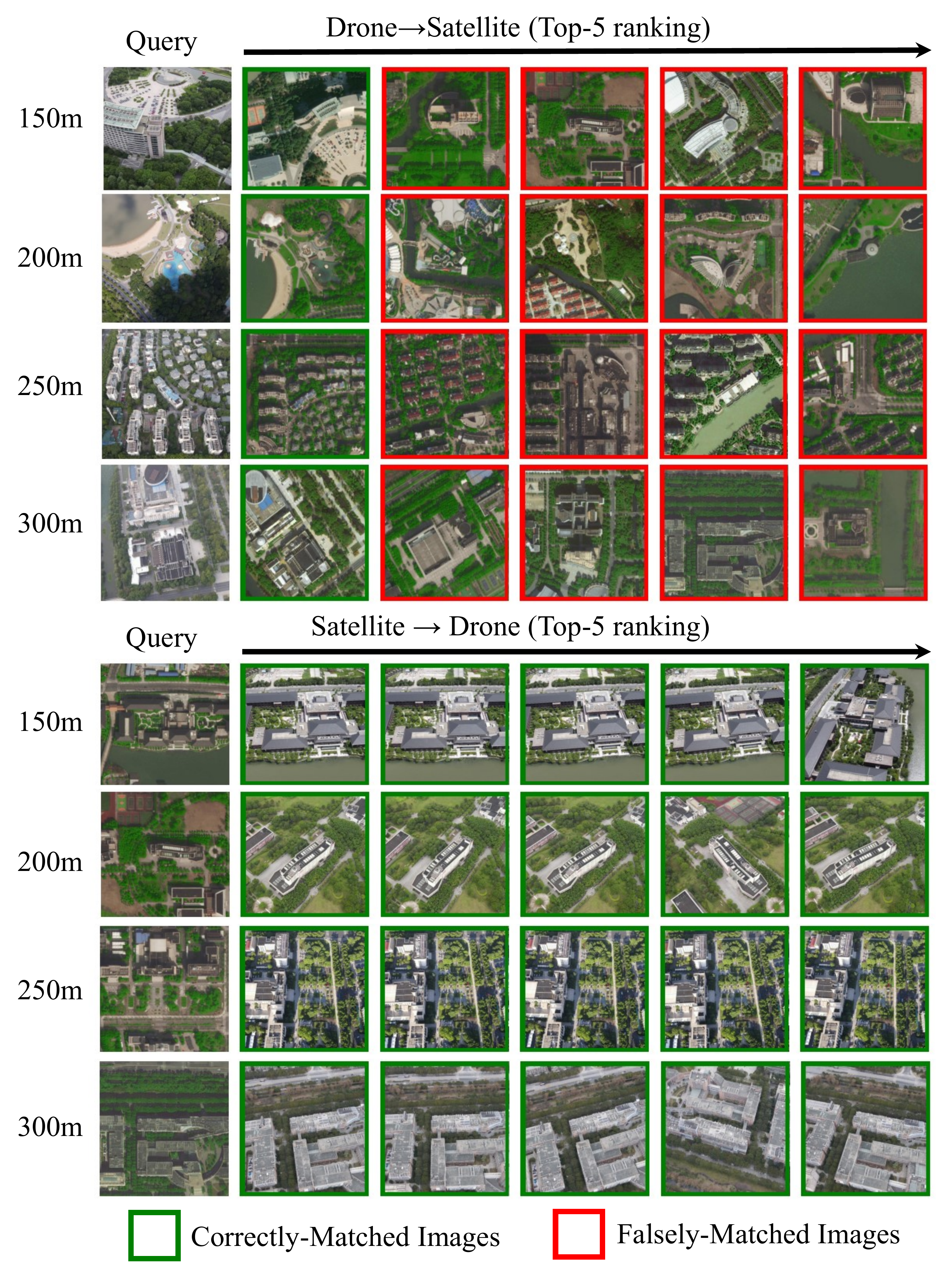}
    \caption{Top-5 Retrieval Results of the Proposed MEAN on the SEUS-200 Dataset.}
    \label{fig6}  
\end{figure}

Fig.~\ref{fig4}(f-j) further illustrates the feature distributions of different models in a 2D feature space obtained using t-SNE \cite{van2008visualizing}. The initial features (Fig.~\ref{fig4}(f)) and the baseline model (Fig.~\ref{fig4}(i)) show poor separation between features from different locations, whereas features from the same location are widely dispersed. In comparison, MCCG (Fig.~\ref{fig4}(g)) and Sample4Geo (Fig.~\ref{fig4}(h)) show some improvements in intra-class compactness and inter-class separation though partial overlap persists between features of different locations. In contrast, the MEAN model (Fig.~\ref{fig4}(j)) demonstrates a distinct advantage: images from the same location, including drone and satellite views, are closely clustered in feature space, while features from different locations are clearly separated. This distribution highlights the strong modality invariance of the MEAN model, effectively mitigating feature discrepancies caused by viewpoint variations and enhancing cross-view consistency.
\subsection{Retrieval Results}
To further illustrate the effectiveness of the MEAN model, we present the retrieval results on the University-1652 dataset as shown in Fig.~\ref{fig5}. For each retrieval result, the green border indicates correctly matched images, while the red border signifies incorrect matches. The results indicate that the MEAN model significantly improves retrieval performance.

In the drone-to-satellite task, the "Query" panel on the left displays the query image from the drone perspective, while the right panel shows the satellite images retrieved by the model. Given that there is only one satellite image per location, MEAN successfully retrieves the correct satellite image in the Top-1 ranking for each scene, demonstrating its accuracy and effectiveness under single-image conditions.

In the satellite-to-drone task, the satellite perspective image serves as the query image, and MEAN efficiently matches the correct target in the drone perspective (green border). This outcome illustrates that MEAN maintains a high level of retrieval performance in CVGL tasks, effectively capturing feature consistency across different viewpoints.

Furthermore, to further validate the performance of MEAN in multiscale environments and under significant viewpoint differences, we visualize the matching performance on the SEUS-200 dataset at various altitudes (150m, 200m, 250m, and 300m) as shown in Fig.~\ref{fig6}. The results indicate that even in scenarios with substantial scale and perspective discrepancies, MEAN maintains more stable modality invariance between different viewpoints and effectively learns modality consistency across multiple height settings.

\section{Discussion}\label{discussion}

 The key difference between our proposed method and previous methods lies in the former's ability to achieve a better trade-off between computational efficiency and model performance. CVGL is primarily deployed in resource-constrained scenarios like UAV platforms, where computational efficiency is of critical importance. State-of-the-art methods, such as CCR \cite{du2024ccr}, Sample4Geo \cite{deuser2023sample4geo}, SRLN \cite{lv2024direction}, and DAC \cite{xia2024enhancing}, often rely on high-performance feature extractors and complex module designs, and typically prioritize performance improvements at the expense of computational efficiency. As a result, they are less suitable for deployment in resource-constrained scenarios. Unlike existing methods that employ multi-head self-attention mechanisms in Transformer-based architectures \cite{liu2021swin} or deep convolutional networks \cite{liu2022convnet} for feature extraction, we adopt a lightweight shallow backbone for initial feature extraction. We present a solution to address the inherent limitations of shallow architectures with suboptimal initial performance in cross-view feature representation. Through multiple embedding generation and fusion, global embedding representation learning, and cross-domain enhanced alignment, our method effectively captures cross-view structural and spatial representations without significantly increasing the parameter scale. 
 
It should be noted that our method achieves excellent performance using fewer parameters. The lightweight 9-layer shallow backbone of a compact design significantly reduces the parameter scale and architectural complexity and achieves a 67\% reduction in the parameter count compared with the methods \cite{zhao2024transfg,deuser2023sample4geo,xia2024enhancing,ge2024multibranch}. An efficient local and global feature enhancement strategy is proposed utilizing two parallel branches instead of the traditional deep modules and the attention mechanism to counteract the efficiency degradation induced by the shallow backbone. These two simple branches significantly enhance the semantic completeness and discrimination of shallow features with a lightweight network architecture. The effectiveness of the two branches benefits from the design of dilation rates in the dilated convolutions. We test different combinations of dilation rates and ultimately select a configuration that obtains the best receptive field coverage and semantic representation capability within the multi-scale structure. In addition, the parallel branches offer flexible optimization space to independently modulate the learning process, which promote more efficient parameter convergence toward the optimal solution of the loss functions. Furthermore, in the third branch, we incorporate a dimensionality reduction strategy at intermediate layers to lower computational expense and preserve the high- and low-dimensional spatial information. Ultimately, our method achieves competitive or superior performance with a lower computational overhead.

\section{CONCLUSION}\label{conclusions}
In this paper, we propose a lightweight MAEN framework for CVGL tasks. The framework aims to enhance feature representation capability and discriminability through a multi-branch structure, comprising progressive diversification embedding, global extension embedding, and cross-domain enhanced alignment. The progressive diversification embedding branch focuses on generating diverse feature embeddings to accommodate complex geographic view variations and utilizing contrastive learning to improve feature consistency and discriminability. The global extension embedding branch further optimizes the interaction between global and fine-grained features to realize the coherent expression of cross-domain information. The cross-domain enhanced alignment branch learns a shared mapping between domains through adaptive calibration and applies a cross-domain invariance alignment loss to overcome the limitations of relying on local detail alignment. This strengthens the intrinsic correlation of cross-domain features and further mines consistency and invariance within the embedding space. Experimental results demonstrate that MAEN achieves competitive performance to strike a notable balance between matching accuracy and computational efficiency, while exhibiting superior adaptability and robustness in CVGL tasks and outperforming existing state-of-the-art methods in certain cases.

In current CVGL tasks, existing methods heavily rely on labeled paired images and label-driven supervised training. In future work, we will explore a new self-supervised learning framework to alleviate the bottleneck of high data annotation costs in CVGL.

\begingroup
\footnotesize
\bibliographystyle{IEEEtran}
\bibliography{reference}
\endgroup
\end{document}